\documentclass[final,5p,times]{elsarticle}

\usepackage{amssymb}
\usepackage{amsmath}  
\usepackage{graphicx} 
\usepackage{caption}
\usepackage{epstopdf}
\usepackage{subfigure}
\usepackage{color}
\usepackage{xcolor}
\usepackage{booktabs}
\usepackage{tabularx}
\usepackage{bm}

\journal{xxx}

\begin{document}

\bibliographystyle{elsarticle-num}
\begin{frontmatter}



\title{Time-Frequency Analysis based Blind Modulation Classification for Multiple-Antenna Systems }

\author[label1]{Weiheng Jiang}  \author[label1]{Xiaogang Wu} \author[label1]{Bolin Chen} \author[label1]{Wenjiang Feng} \author[label2]{Yi Jin}

\address[label1]{School of Microelectronics and Communication Engineering, Chongqing University, Chongqing 400044, China.}
\address[label2]{Xi'an Branch of China Academy of Space Technology, Xi'an 710100, China.}

\begin{abstract}
Blind modulation classification is an important step to implement cognitive radio networks. The multiple-input multiple-output (MIMO) technique is widely used in military and civil communication systems. Due to the lack of prior information about channel parameters and the overlapping of signals in the MIMO systems, the traditional likelihood-based and feature-based approaches cannot be applied in these scenarios directly. Hence, in this paper, to resolve the problem of blind modulation classification in MIMO systems, the time-frequency analysis method based on the windowed short-time Fourier transform is used to analyse the time-frequency characteristics of time-domain modulated signals. Then the extracted time-frequency characteristics are converted into RGB spectrogram images, and the convolutional neural network based on transfer learning is applied to classify the modulation types according to the RGB spectrogram images. Finally, a decision fusion module is used to fuse the classification results of all the receive antennas. Through simulations, we analyse the classification performance at different signal-to-noise ratios (SNRs), the results indicate that, for the single-input single-output (SISO) network, our proposed scheme can achieve 92.37\% and 99.12\% average classification accuracy at SNRs of -4 dB and 10 dB, respectively. For the MIMO network, our scheme achieves 80.42\% and 87.92\%  average classification accuracy at -4 dB and 10 dB, respectively. This outperforms the existing classification methods based on baseband signals.
\end{abstract}


\end{frontmatter}

\section{Introduction}
The increase in communication demands and the shortage of spectrum resources has caused the cognitive radio (CR) and multiple-input multiple-output (MIMO) techniques to be implemented in wireless communication systems. As one of the essential steps of CR, modulation classification (MC) is widely applied in both civil and military applications, such as spectrum surveillance, electronic surveillance, electronic warfare, and network control and management \cite{8471222}. It improves radio spectrum utilisation and enables intelligent decision-making for context-aware autonomous wireless spectrum monitoring systems \cite{liao2019sequential}. However, most of the existing MC methods are focussed  on single-input single-output (SISO) scenarios, which cannot be directly applied when multiple transmit antennas are equipped at the transceivers \cite{7384691}. Therefore, it is crucial to research the performance of the MC method for MIMO communication systems.

Traditional MC approaches for the SISO systems discussed in the literature can be classified into two main categories: likelihood-based (LB) approaches and feature-based (FB) approaches \cite{4167652}. The LB approaches can theoretically achieve optimal performance as they compute the likelihood functions of the different modulated signals to maximise the classification accuracy. However, they have  a very high computational complexity and require prior information, such as the channel coefficient \cite{5461606} \cite{8333735}. Hence, the LB approaches cannot  be directly applied in fast modulation classification and blind modulation classification (BMC). By contrast, the FB approaches cannot obtain the optimal result, but they have lower computational complexity and do not require prior information. The FB methods usually include two steps:  feature extraction and classifier design. The higher-order statistics, instantaneous statistics,  and other features are calculated in the feature extraction. Then  the popular classification methods, such as decision tree \cite{7295481}, support vector machine \cite{7728143} \cite{8017570}, and artificial neural network (ANN) \cite{8754798} \cite{UAV} are adopted as the classifiers.

With the rapid rise of artificial intelligence and the emerging requirements of intelligent wireless communication, deep learning-based approaches are now becoming widely studied and used in different aspects of wireless communication, such as the transceiver design at the physical layer \cite{8054694} and BMC problems \cite{ramjee2019fast} \cite{8357902} \cite{8643801} \cite{8760481} \cite{8454504} \cite{8761426}. As for BMC in SISO scenarios, the raw in-phase and quadrature phase (IQ) data or the time-domain amplitude and phase data can be directly used as the input of the deep learning neural network. More specifically, the authors in \cite{ramjee2019fast} presented convolutional long short-term deep neural network and deep residual network (Resnet) algorithms to identify 10 different modulation types, with a high classification accuracy over a wide range of signal-to-noise ratio (SNR) values. Rajendran et al. \cite{8357902} proposed a new data-driven model for BMC based on long short-term memory (LSTM), which learnt the features from the time-domain amplitude and phase information of the modulation schemes and yielded an average classification accuracy close to 90\% for SNRs from 0 dB to 20 dB. Zhang et al. \cite{8643801}, adopting the Resnet model as the classifier, presented an approach to fuse the time-frequency images and the handcrafted features of the modulated signals to obtain more discriminating features. The experimental results showed that the proposed scheme has a superior performance. The latest research indicates that the deep learning-based MC methods achieve better overall performance than the traditional LB and FB approaches for the SISO systems.

Although the MC (or BMC) method systems on SISO networks are becoming more mature, research into using MC for MIMO networks has just begun \cite{7384691}. The authors in \cite{6117042} and \cite{7459788} proposed similar methods for the MC of MIMO transceiver systems, which calculate the higher order statistical moments and cumulants of the received signal. Then the artificial neural network is employed to classify the modulation types. In \cite{8422500}, a clustering classifier based on centroid reconstruction is presented to identify the modulation scheme with unknown channel matrix and noise variance in MIMO systems. The  simulation results showed that their algorithm can obtain excellent performance, even at low SNR and with a very short observation interval. To deal with the BMC problem and the two major constraints in the railway transmission environment (i.e. the high speeds and impulsive nature of the noise), Kharbech et al. \cite{8356664} proposed a feature-based process of blind identification that includes three parts: impulsive noise mitigation, feature extraction, and classification. By analysing the correlation functions of the received signals for certain modulation formats, Mohamed et al. resolved the BMC problem in single and multiple-antenna systems operating over frequency-selective channels in \cite{MareyBlindMIMO}, and the BMC problem in Alamouti STBC System \cite{MareyBlind}.

For MIMO systems, it is difficult to directly apply deep learning to the raw IQ data or the time-domain amplitude and phase data, since the overlapped signals at the receiver of the MIMO system destroy the statistical features \cite{8533632}. Hence, it is crucial to extract the distinguishable features or convert the raw signals for BMC in MIMO systems. Due to the time-frequency analysis methods can jointly analyse the time-domain and frequency-domain features of signals, and the different modulation types have distinct time-domain and frequency-domain features. Hence, in this paper, in order to overcome the effect of the overlapped signals at the receiver, we analyse the time-frequency features of the modulated signals to resolve the BMC problem in MIMO systems. First, the time-frequency analysis method based on the windowed short-time Fourier transform (STFT) \cite{2016xli} is employed to generate the spectrum of the MIMO-modulated signals. Then the spectrum in different time windows is converted to a greyscale image, and the greyscale image is converted to a red-green-blue (RGB) spectrogram image \cite{OZER2018505}. Second, a fine-tuned AlexNet-based convolutional neural network (CNN) model is introduced to learn the features from the RGB spectrogram images. The modulation scheme of each receiving stream among the receiving MIMO signals is identified in this stage. Finally, the previously produced decisions are merged to form the final result. In addition, this method can be simplified to directly apply to SISO systems. The simulation results show that the proposed method achieves a superior performance at low SNR scenarios for both MIMO and SISO systems.

This paper is organised as follows. The signal model of the MIMO and SISO systems and the STFT-based time-frequency analysis method are introduced in Section \ref{Model_TFD}. Section \ref{Scheme} presents the BMC scheme for MIMO systems, including the proposed CNN model and the decision method. Then the RGB spectrogram image and the classification performance in different scenarios are analysed in Section \ref{Preforemance}. Finally, conclusions are drawn in Section \ref{conclusion}.

\section{Signal model and time-frequency analysis method}
\label{Model_TFD}
In this section, we define the MIMO signal model, and then the simplified SISO signal model is derived. Then the STFT-based time-frequency analysis method is introduced to generate the spectrogram image of the MIMO modulated signals.
\subsection{MIMO signal model }
\label{MIMOModel}
We consider a MIMO-based single-carrier wireless communication system with $N_t$ transmit antennas and $N_r$ receive antennas. The flat-fading and time-invariant MIMO channel is adopted herein. Therefore, the MIMO channel $\mathbf{H} \in \mathbb{C}^{N_r \times N_t}$ is defined as
 \begin{equation}
  \mathbf{H} = \left[ \begin{array}{ccc}
h_{11} & \cdots & h_{1N_t}\\
\vdots & \ddots & \vdots\\
h_{N_r1} & \cdots & h_{N_rN_t}\\
\end{array}
\right],
  \label{MIMOChannel}
 \end{equation}
 where $h_{ij}$ represents the channel coefficient between the $j$-th transmit antenna and $i$-th receive antenna. The channel matrix $\mathbf{H}$ is assumed to be full-column rank and the channel gains remain constant over the observation interval.
 Let $\bm{x} = [x_1(t),...,$ $x_j(t),...,X_{N_t}(t)]^T$ denote the transmitted data streams, where $x_j(t)$ represent the transmitted modulated signal at the $j$th transmit antenna. Likewise, let $\bm{y} = [y_1(t),...,y_i(t),...,y_{N_r}(t)]^T$ represent the received data streams, where $y_i(t)$ is the received signal at the $i$th receive antenna. Then the received signals can be further described by
\begin{equation}
  \bm{y} = \mathbf{H}\bm{x} + \bm{n},
  \label{rec}
 \end{equation}
where the vector $\bm{n} = [n_1(t),...,n_i(t),...,n_{N_r}(t)]^T$ represents the additive white Gaussian noise (AWGN) vector and each element $n_i(t)$ of $\bm{n}$ is an identically and independently distributed (i.i.d.) random variable with zero mean and variance $\sigma^2$ (i.e. $n_i(t) \sim \mathcal{N}(0,\sigma^2)$). In order to obtain the RGB spectrogram image of $y_i(t)$, the datasets generated in this paper are time-domain signals \cite{8643801}, instead of the baseband signals used in \cite{8364579,8846691}.

\subsection{SISO signal model}
\label{SISOModel}
When $N_r=N_t=1$, the MIMO-based signal model in Section \ref{MIMOModel} can be converted into a SISO-based signal model. The received signals corrupted by the AWGN in the SISO system can then be represented as
\begin{equation}
  y(t) = hx(t) + n(t),
  \label{recSISO}
 \end{equation}
where $x(t)$ represents the original digital modulated signals, $y(t)$ represents the digital modulated signals over the wireless channel, $h$ represents the channel attenuation coefficient, and $n(t)$ denotes the AWGN.  In this paper, the original digital modulated signals $x(t)$ may be multiple amplitude-shift keying (MASK), multiple frequency-shift keying (MFSK), multiple phase-shift keying (MPSK) and quadrature amplitude modulation (QAM) signals \cite{proakis2001digital}.
\begin{figure*}[htb]
\centerline{\includegraphics[scale=1.6]{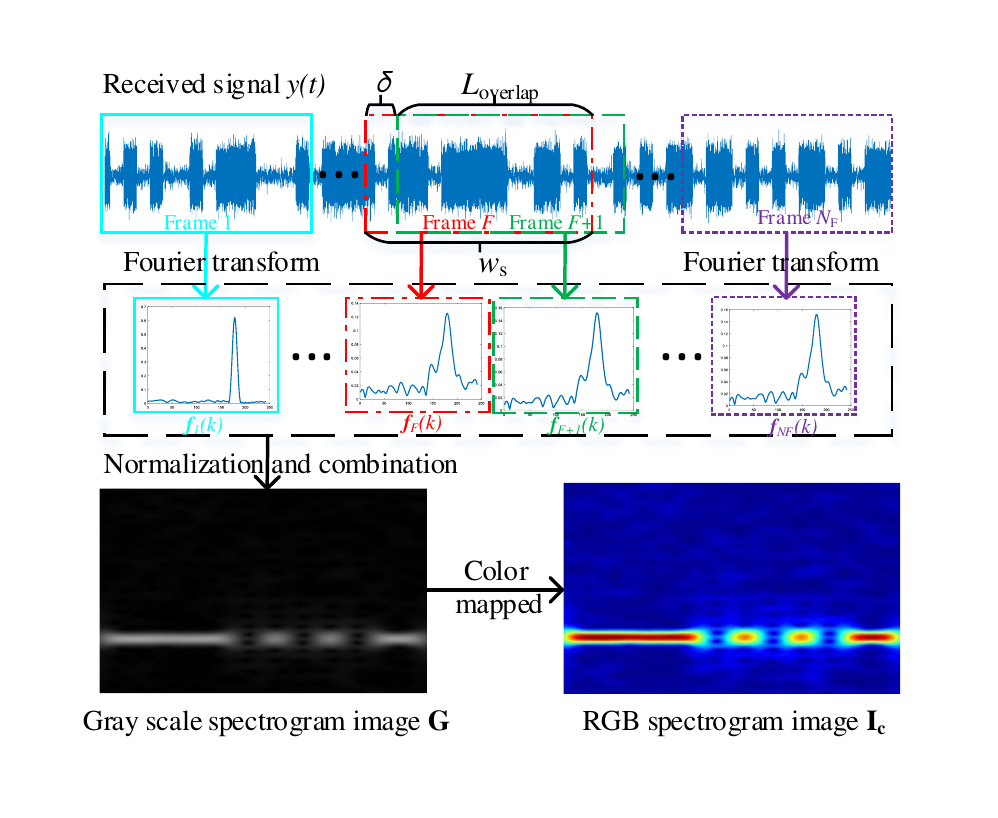}}
\caption{$~$The flow chart of STFT-based time-frequency analysis.}
\label{TFD}
\end{figure*}

The time-domain expression of MASK-modulated signals is described as
\begin{equation}
  x(t) =\sum\limits_{n} A_mg(t-nT_s)cos(2\pi f_ct+\varphi_0),
  \label{MASK}
\end{equation}
where $A_m$, $T_s$, $f_c$, and $\varphi_0$ represent the modulation amplitude, symbol period, carrier frequency, and initial phase, respectively. The value of $A_m$  depends on the symbol sequence and the modulation order $M$. In addition, $g(t)$ is a baseband signal waveform and is usually a square-root raised cosine pulses.

Similarly, the time-domain expressions of MFSK and MPSK are defined as
\begin{equation}
  x(t) = \sum\limits_{n}g(t-nT_s)cos(2\pi f_mt+\varphi_0)
  \label{MFSK}
\end{equation}
and
\begin{equation}
  x(t) = \sum\limits_{n}g(t-nT_s)cos(2\pi f_ct+\varphi_m+\varphi_0),
  \label{MPSK}
\end{equation}
respectively.

In (\ref{MFSK}) and (\ref{MPSK}), $f_m$ and $\varphi_m$ are the modulation frequency and phase, respectively, and the values of these parameters depend on the symbol sequence and the modulation order $M$.

However, the QAM signal is slightly different from the MXSK (MASK, MFSK, and MPSK) modulated signals, because the QAM-modulated signal has two orthogonal carriers. Therefore, it can be represented as
\begin{equation}
    \begin{aligned}
            x(t) &= \sum\limits_{n}a_ng(t-nT_s)cos(2\pi f_ct+\varphi_0)\\
              &+ \sum\limits_{n}b_ng(t-nT_s)sin(2\pi f_ct+\varphi_0),
    \end{aligned}
  \label{MQAM}
\end{equation}
where $a_n, b_n \in [2m-1-\sqrt{M}]$, $m = 1,2,...,\sqrt{M}$, and the two carriers are individually modulated by $a_n$ and $b_n$ \cite{8017570}.

\subsection{STFT-based time-frequency analysis}
\label{STFT}\
In this paper, the STFT  is adopted in the modulated signal analysis. That is, we use STFT to analyse the frequency and phase of local sections of the time-varying modulated signals with a time window function \cite{7181637}. Then the spectrogram image (the visual representation of the frequency spectrum of a signal)  is constructed. In this subsection, we introduce the theory of the STFT, and then we present the method to generate the STFT-based RGB spectrogram image for the modulated signals.

\subsubsection{Theory of the STFT}
Consider a signal $s(\tau)$ and a real, even window $w(\tau)$, whose Fourier transforms (FT) are $S(f)$ and $W(f)$, respectively. To obtain a localised  spectrum of $s(\tau)$ at time $\tau = t$, the signal is multiplied by the window $w(\tau)$ centred at time $\tau = t$, which results in
\begin{equation}
  s_w(t,\tau) = s(\tau)w(\tau-t).
  \label{windSignal}
\end{equation}
Next, the FT at is taken at time $\tau$, obtaining
\begin{equation}
   \mathbb{F}_s^w(t,f)= \mathop{\mathcal{F}}\limits_{\tau \rightarrow f}s(\tau)w(\tau-t),
  \label{windSignal}
\end{equation}
where $\mathbb{F}_s^w(t,f)$ is the STFT \cite{2016xli}.

\subsubsection{Generating the STFT-based RGB spectrogram image for the modulated signals}
In order to perform the STFT and obtain the spectrogram image of the modulated signals, we implement the process showed in Fig. \ref{TFD}. Dividing a given discrete modulated signal vector $y(n)$ of length $L$ into highly overlapped frames each with length $w_s$ generates the spectral vector $\bm{f}$, where $y(n)$ is obtained by sampling the received modulated signal $y(t)$. Hence, the signal in the current frame, $y_F(n)$, is
\begin{equation}
  y_F(n) = y(F\delta + n)w(n), n = 0,...,w_s-1,
  \label{Signaloverlap}
\end{equation}
where $F$ is the current frame, $w(n)$ is the window function, the window function can be hamming,hanning or blackman, and we choose hamming in this paper \cite{mitra2006digital}. Then the $\delta$ is the incrementation between two consecutive frames, which is calculated by
\begin{equation}
  \delta = w_s - L_{overlap}.
  \label{delta}
\end{equation}

Herein, the $L_{overlap}$ ($L_{overlap} < w_s <L$) is the length of overlapped signals between two consecutive frames, and the number of frames $N_F$ can be calculated by
\begin{equation}
  N_F = \frac{L-L_{overlap}}{\delta} = \frac{L-L_{overlap}}{w_s-L_{overlap}}.
  \label{NumofFrame}
\end{equation}
The larger the $L_{overlap}$, the greater the $N_F$, and hence the higher time resolution of the STFT.

The hamming window function $w(n)$ is defined as
\begin{equation}
  w(n)= \left[0.54-0.46cos\left(\frac{2\pi n}{w_s-1}\right)\right]R_{w_s}(n),
  \label{hamming}
\end{equation}
where $R_{w_s}(n)$ is a rectangular window with length $w_s$.

Based on (\ref{Signaloverlap}), we can obtain the spectral magnitude vector $\bm{f}_F$ of the current frame $F$,
\begin{equation}
  \bm{f}_F(k) = \sum\limits_{n=0}^{w_s-1}y_F(n)e^{-j2\pi nk/N}, k = 1,...,(N/2 -1)
  \label{spectral}
\end{equation}
where $N/2-1$ is the number of points of the Fourier transform. The larger the $N$, the higher the frequency resolution of the STFT. Therefore, the linear value of the spectral magnitude vector  is obtained as
\begin{equation}
  \bm{S}(k,F) = |\bm{f}_F(k)|.
  \label{LinerSpectral}
\end{equation}

The linear value of the spectral magnitude vector can be normalised in the range of [0, 1] as
\begin{equation}
  \bm{G}(k,F) = \frac{\bm{S}(k,F)-min(\bm{S})}{max(\bm{S})-min(\bm{S})}.
  \label{NorSpec}
\end{equation}

By combining the normalised linear spectral magnitude vector $\bm{G}(k,F)$ of all the frames as
\begin{equation}
  \mathbf{G} = [\bm{G}(k,1)^T;...;\bm{G}(k,F)^T;...;\bm{G}(k,N_F)^T],
  \label{matrix}
\end{equation}
we can obtain the time-frequency matrix $\mathbf{G}\in \mathbb{C}^{(N/2-1) \times N_F}$. This matrix is a greyscale image of the spectral magnitude vector, the size of this image is $(N/2-1) \times N_F$, the horizontal axis of this image represents time, and the vertical axis represents frequency.

Next, the greyscale image is quantised into its RGB components, the mapping type is the $jet$ in matlab r2016b \cite{jet}. The mapping is expressed as
\begin{equation}
  \mathbf{I}_c = f_{map}(\mathbf{G}),
  \label{map}
\end{equation}
where $\mathbf{I}_c$ is the RGB spectrogram image and $f_{map}$ is the non-linear $jet$ quantisation function \cite{OZER2018505}. It is worth noting that, to facilitate the observation and analysis of RGB spectrogram image, we deploy the color mapping in this paper, this step can be omitted in practical applications.

For the STFT, by adjusting the values of the window length $w_s$ and overlapped signal length $L_{overlap}$, we can tune the time resolution of the RGB spectrogram image. Moreover, by adjusting the number of points of the Fourier transform $N/2-1$,  we can also tune the frequency resolution of the RGB spectrogram image.

\begin{figure}[htbp]
\centerline{\includegraphics[scale=0.7]{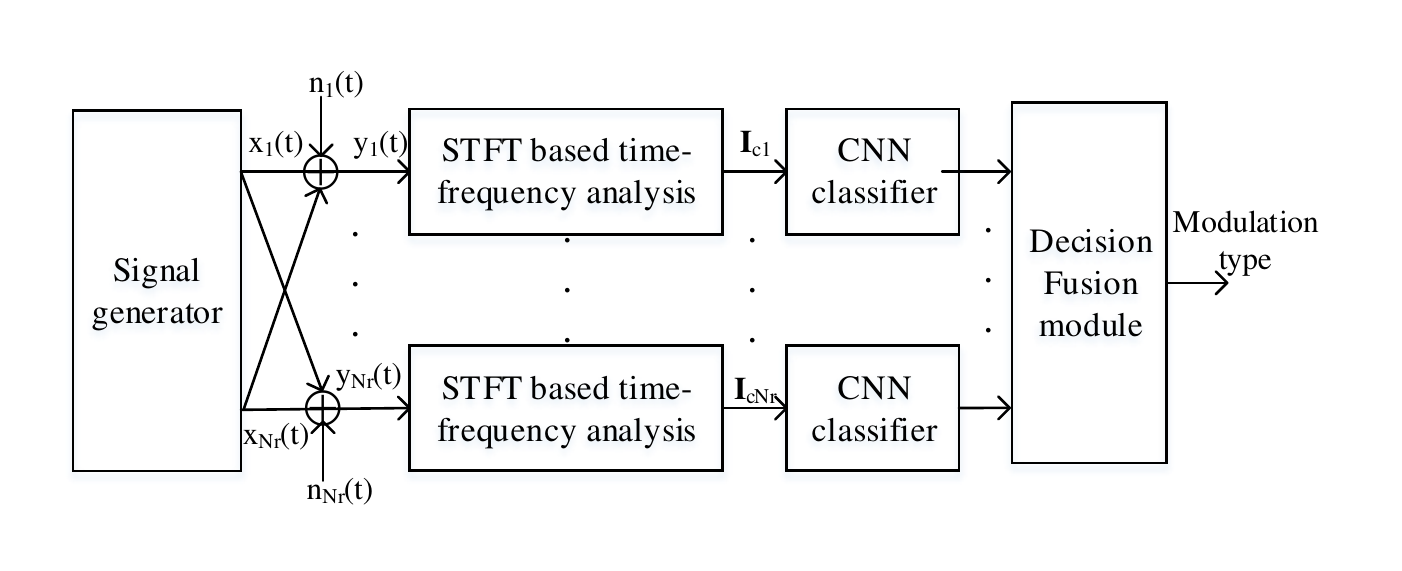}}
\caption{$~$Block diagram of the proposed MIMO modulation classification scheme.}
\label{MIMOmodulation}
\end{figure}

\section{Proposed BMC scheme}
\label{Scheme}
In this section, a time-frequency analysis is conducted and a deep learning-based BMC scheme is proposed. The block diagram of the proposed BMC scheme is shown in Fig. \ref{MIMOmodulation}, which shows four modules: signal generator, time-frequency analysis, CNN classifier, and decision fusion. The signal generator outputs the modulated signals $x_i(t)$ (with the same modulation type) for each transmit antenna \cite{6117042}. This process was described in subsections \ref{MIMOModel} and \ref{SISOModel}. Then the time-frequency analysis is performed for the received signal $y_i(t)$ for each receive antenna, which generates the RGB spectrogram image $\mathbf{I}_{ci}$ (partially described in subsection \ref{STFT}). Next, the AlexNet-based CNN classifier is trained based on a number of RGB spectrogram images in the training stage, and the modulation type of each received signal $y_i(t)$ is identified in the test stage. Finally, the decisions of different signal branches are combined by the decision fusion module for the final decision. In the next three sections , we will illustrate in detail the procedures of the time-frequency analysis, CNN based classifier, and decision fusion.

\begin{figure*}[htbp]
\centerline{\includegraphics[scale=1]{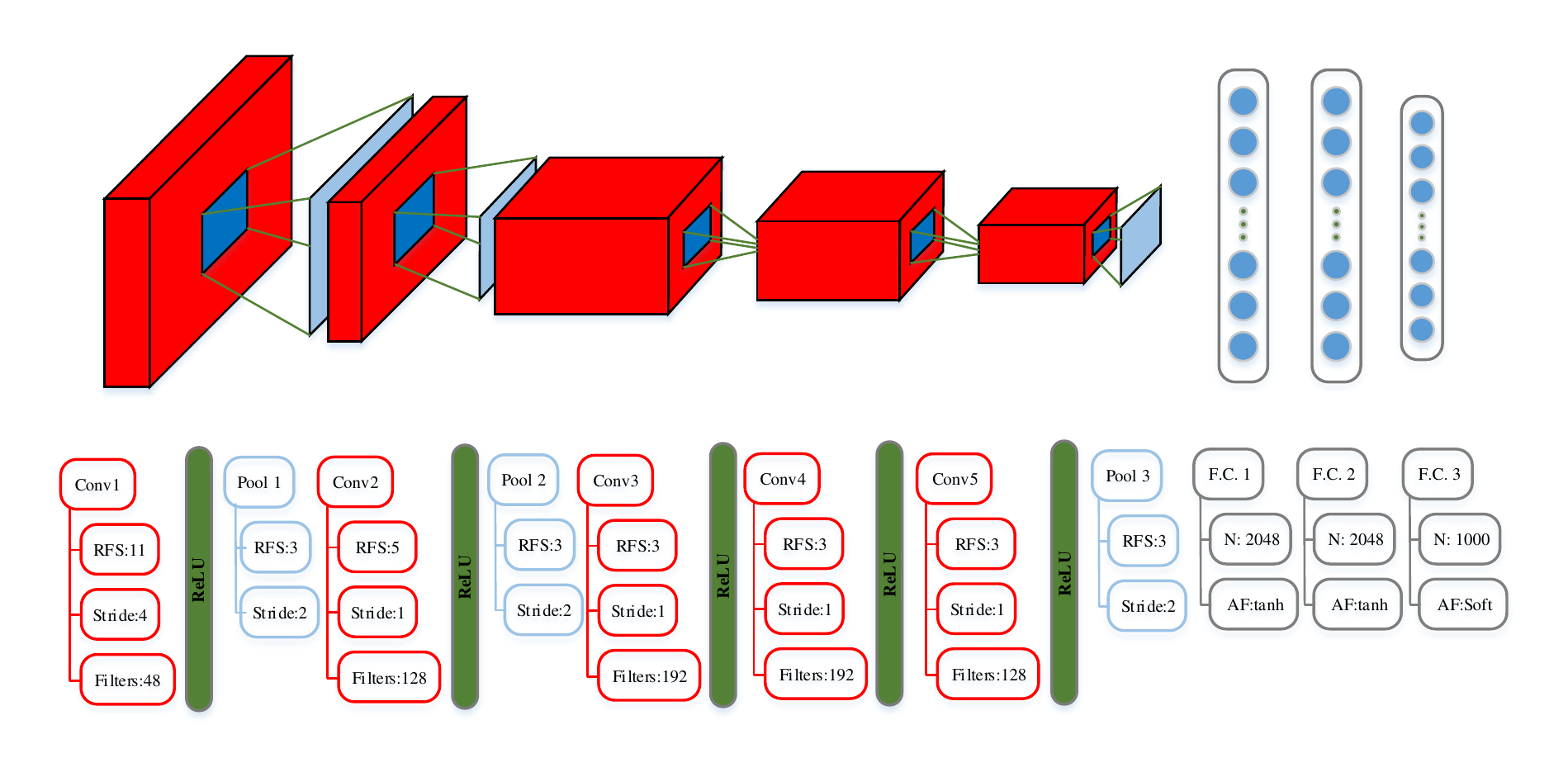}}
\caption{$~$Architecture of AlexNet (Conv: convolution layer, Pool: pooling layer, F.C.: fully connected layer, RFS: receptive field size, N: number of neurons in fully connected layer, AF: activation function, Soft: Softmax) \cite{8401505}.}
\label{AlxNetFigure}
\end{figure*}

\subsection{Time-frequency analysis for received signals}
\label{imageFeature}
The flow chart of STFT-based time-frequency analysis is shown in Fig. \ref{TFD}. First, using the ASK signal as an example, the received signal $y(t)$ is divided into $N_F$ frames by the hamming window $w(n)$ with length $w_s$, the details of which are described in Eqs.  (\ref{Signaloverlap})-(\ref{hamming}). Second, the spectrum of the windowed signal is obtained by its Fourier transform. Third, by normalising and combining the linear spectral magnitude vector, the greyscale spectrogram image $\mathbf{G}$ is obtained (the size of the related greyscale matrix is $(N/2-1) \times N_F$). Finally, to accommodate the input layer of AlexNet and improve the distinguishability  of the spectrogram image, the greyscale spectrogram image is mapped into RGB spectrogram image $\mathbf{I}_c$ (the size of the related RGB matrix is $(N/2-1) \times N_F \times 3$). Then, the RGB matrix is cut or padded into $227 \times 227 \times 3$ before feeding it into the CNN.

\subsection{AlexNet based CNN classifier}
\label{AlexNetCNN}
In our proposed BMC scheme, AlexNet, which is utilised for object detection \cite{Krizhevsky2012ImageNet} and was the winner of the 2012 ImageNet Large Scale Visual Recognition Challenge (ILSVRC), is adopted as the classifier. The network architecture of AlexNet is shown in Fig. \ref{AlxNetFigure} \cite{8401505}.

As depicted in Fig. \ref{AlxNetFigure}, AlexNet contains eight layers; the first five are convolutional and the remaining three are fully connected . The output of the last fully connected layer is fed to a 1000-way softmax that produces a distribution over the 1000 class labels \cite{Krizhevsky2012ImageNet}. AlexNet uses the rectified linear unit (ReLU) as the activation function of the CNN. In practice, the dropout and max pooling techniques are applied to the CNN. AlexNet has an excellent performance in visual tracking and object detection due to its capability in  sensing the pattern position on the image. Therefore, considering that the spectrogram image has rich pattern position information, it is  sensible to choose AlexNet as the classifier network.

The motivation of transfer learning comes from the fact that people can intelligently apply knowledge learned previously to solve new problems faster or with better solutions \cite{5288526}. In order to utilise the pretrained AlexNet, transfer learning is employed to fine-tune AlexNet and accelerate the training process. The last layer of the pretrained AlexNet network in Fig. \ref{AlxNetFigure} is configured with 1000 classes, and this layer must be fine-tuned to accommodate the new classification task. First, all layers except the last layer are extracted, then the last layer is replaced with a new fully connected layer that contains eight  neurons (i.e. the number of modulation categories in this paper). In the end, the parameters of the activation layer and the classification output layer are set to accommodate the new classification task. Therefore, with such fine-tuning, the output of AlexNet can precisely perform the modulation classification of the received signals.

\subsection{Decision fusion}
\label{Decision}
Since there are multiple antennas at the receiver of the MIMO network, it is possible for each branch to cooperate with each other to achieve higher identification reliability \cite{6117042}. As shown in Fig. \ref{MIMOmodulation}, the $N_r$ received signals are classified independently because the influences of signal overlapping, interchannel noise, and random phase shifting may cause each received signal to be identified as a different modulation type. This may lead to incorrect identification results. The decision fusion among all the receive antennas aims to improve the average classification accuracy. The decision vector of the $i$-th received signal, $\bm{d}_i$,  can be defined as
\begin{equation}
  \bm{d}_i = [d_{i1},...,d_{ik},...,d_{iK}],
  \label{dicisionv}
\end{equation}
where $K$ is the number of modulation types, $d_{ik}$ is the probability of identifying the received signal $y_i(t)$ as modulation type $k$, and $d_{ik}$ meets the following condition,
\begin{equation}
  \sum\limits_{k=1}^K d_{ik} = 1.
  \label{dicisionc}
\end{equation}
Therefore, the modulation type $m_i$ of the received signal $y_i(t)$ is the modulation type which has the maximum probability. The modulation type with maximum probability can be defined as a set $\mathcal{M}$ as follows,

\begin{equation}
  \mathcal{M} = \mathop{argmax}\limits_{k\in\{1,...,K\}}d_{ik},
  \label{dicisionc}
\end{equation}
Note that there are two cases for the above equations, 1) the maximum probability is unique, i.e., $|\mathcal{M}| =1$, the modulation type of $i$-th received signal is the element of $\mathcal{M}$; 2) the maximum probability is not unique, i.e., $|\mathcal{M}| \geq 2$, the modulation type of $i$-th received signal is randomly chosen from $\mathcal{M}$.

Hence, the decision fusion can be converted to the problem of deciding the final modulation type $m$ according to $m_i$, $i = 1,2,...,N_r$. The fusion rule at the fusion module can be OR, AND, or majority rule, which can be generalised as the \emph{``n-out-of-$N_r$ rule''}. \cite{Atapattu2014Energy}. That is, a certain modulation scheme is identified when a classifier is decided on among the $N_r$ classifiers. Take the $N_r = 4$ as an example and the possible modulation types formulate the set $\mathcal{M}=$\{2PSK, 4PSK, 8PSK\}, if there are more than three classifiers identify the modulation type as 2PSK (4PSK or 8PSK), then the final modulation type is 2PSK (4PSK or 8PSK); if there are two classifiers identify the modulation type as 2PSK and the other two classifiers identify the modulation type as 4PSK and 8PSK, respectively, then the final decision is 2PSK; in addition, if the two classifiers identify the modulation type as 2PSK and the other two classifiers identify the modulation type as 4PSK (or 8PSK), the decision fusion centre will randomly choose a modulation type between 2PSK and 4PSK (or 8PSK) as the final result.

\section{Performance analysis}
\label{Preforemance}
In this section, the proposed time-frequency analysis and deep learning-based BMC algorithm is tested under different modulation schemes in both the SISO and MIMO scenarios. Specifically, the random channel attenuation assigns a value from $[0,1]$, and random phase shifts within one symbol interval are considered for the MIMO scenario. The  AWGNs with different SNRs are added into the modulated signals for both the SISO and MIMO scenarios. In addition, we consider the following MIMO antenna configurations: $N_t = 2$ and $N_r = 4$. In the simulations, the 2ASK, 2FSK, 2PSK, 4ASK, 4FSK, 4PSK, 8PSK, and 16QAM modulation schemes are considered, unless otherwise stated. The parameters of the modulated signals are assigned as follows. The sampling frequency $f_s$ is 16 KHz, the carrier frequency $f_c$ is 2 KHz, the symbol rate $f_b$ is 100 Hz, and the length of original digital signal is 14 (i.e. each modulated signal contains $(16000/100) \times 14 = 2240$ sample points). In addition, in the training stage, 100 modulated signals for each modulation type and SNR are randomly generated for both the SISO and MIMO scenarios, in which the SNR varies from -4 to 10 dB at intervals of 2 dB \cite{8643801}. In the test stage, 100 modulated signals for each modulation type and SNR are randomly generated. All the signal samples are generated in Matlab 2017b, and the training and testing of AlexNet are based on the Matlab neural network toolbox. Additionally, the parameters to generate the RGB spectrogram image are set as $w_s = 320$, $L_{overlap} = 315$, $\delta = 5$, and $N = 2048$.

We now discuss how the modulation order, SNR, and overlapping of the MIMO signals influence the RGB spectrogram image of the modulated signals. Then the classification performance of the proposed scheme is validated for different scenarios.
\subsection{RGB spectrogram image of the modulated signals}
\label{RGBDiscuss}
\begin{figure}
\centering
\subfigure[the binary signal]{
\begin{minipage}[b]{0.5\textwidth}
\includegraphics[width=\textwidth]{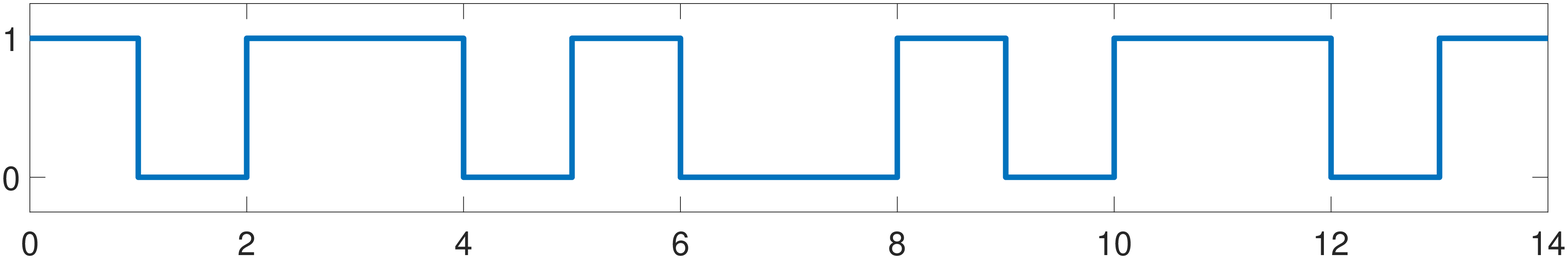} \\
\label{sequencea}
\end{minipage}
}
\subfigure[the quanternary signal]{
\begin{minipage}[b]{0.5\textwidth}
\includegraphics[width=\textwidth]{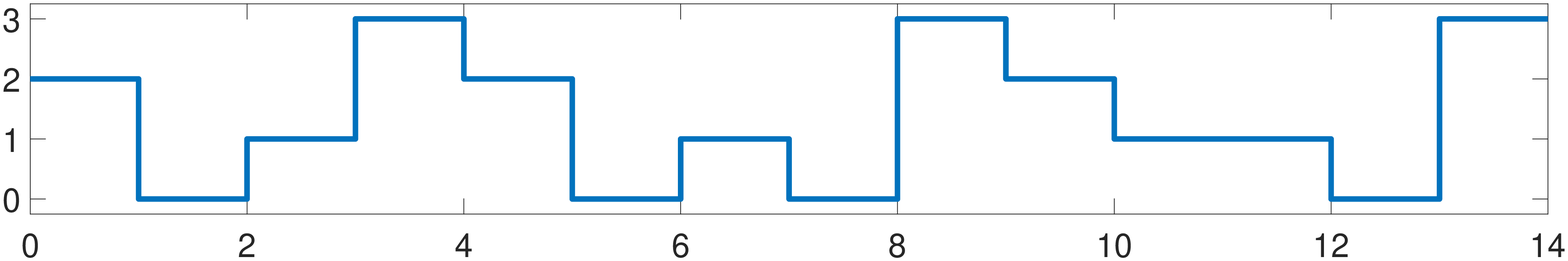} \\
\label{sequenceb}
\end{minipage}
}
\caption{$~$Digital signal sequence before modulation.}
\label{sequence}
\end{figure}
In this subsection, in order to simplify the analysis, we select only certain binary and quaternary digital signal sequences (as shown in Fig. \ref{sequence}) to generate the RGB spectrogram image. The binary signal \ref{sequencea} is used to generate the 2-order modulated signals (i.e. 2ASK, 2FSK, and 2PSK) and the quaternary signal \ref{sequenceb} is used for the 4-order modulated signals (i.e. 4ASK, 4FSK, and 4PSK).
\subsubsection{RGB spectrogram image of the modulated signals with different modulation orders}
\begin{figure*}
\centering
\subfigure[2ASK]{
\begin{minipage}[b]{0.31\textwidth}
\includegraphics[width=\textwidth]{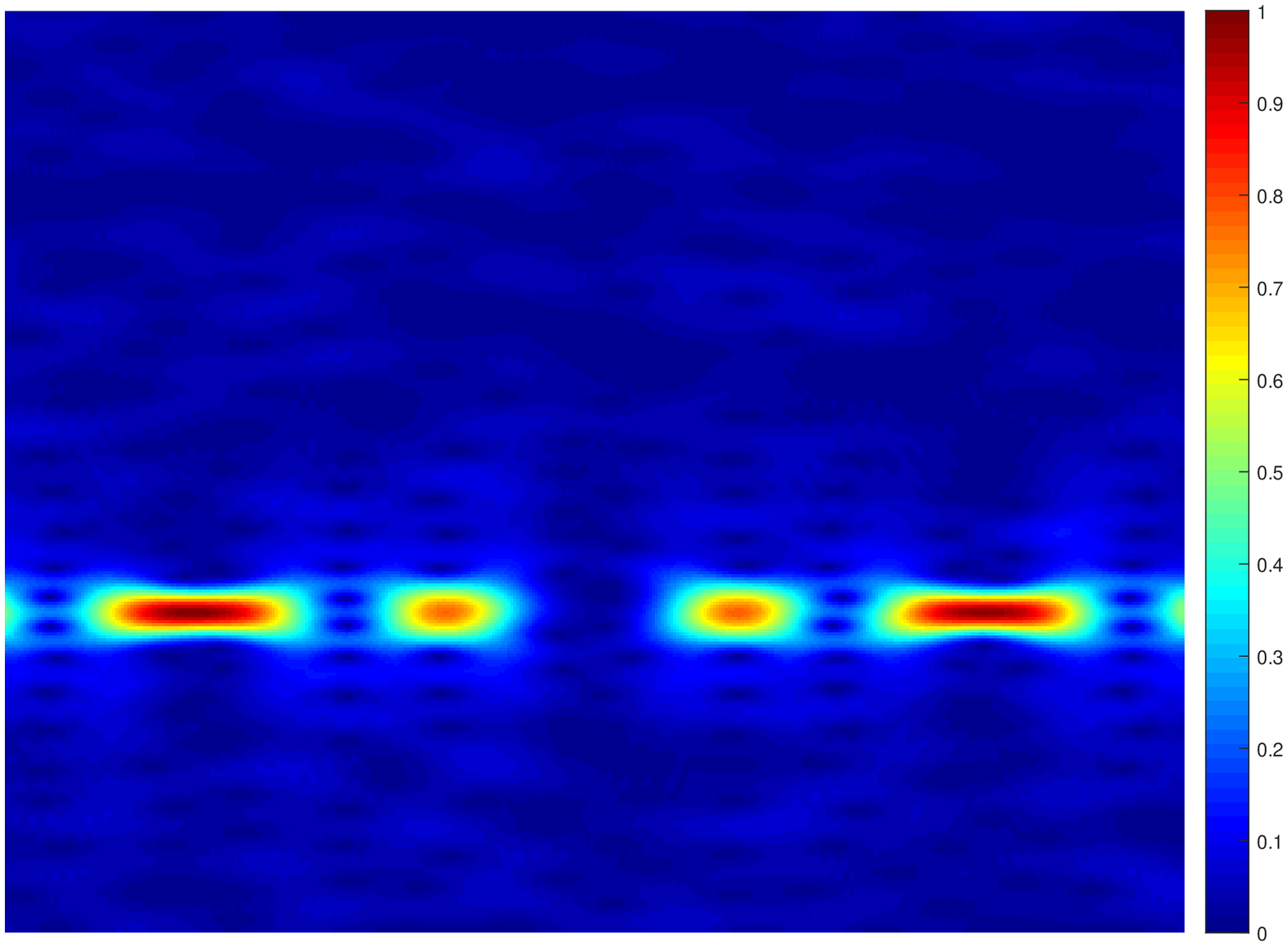} \\
\label{2ASK10dB}
\end{minipage}
}
\subfigure[2FSK]{
\begin{minipage}[b]{0.31\textwidth}
\includegraphics[width=\textwidth]{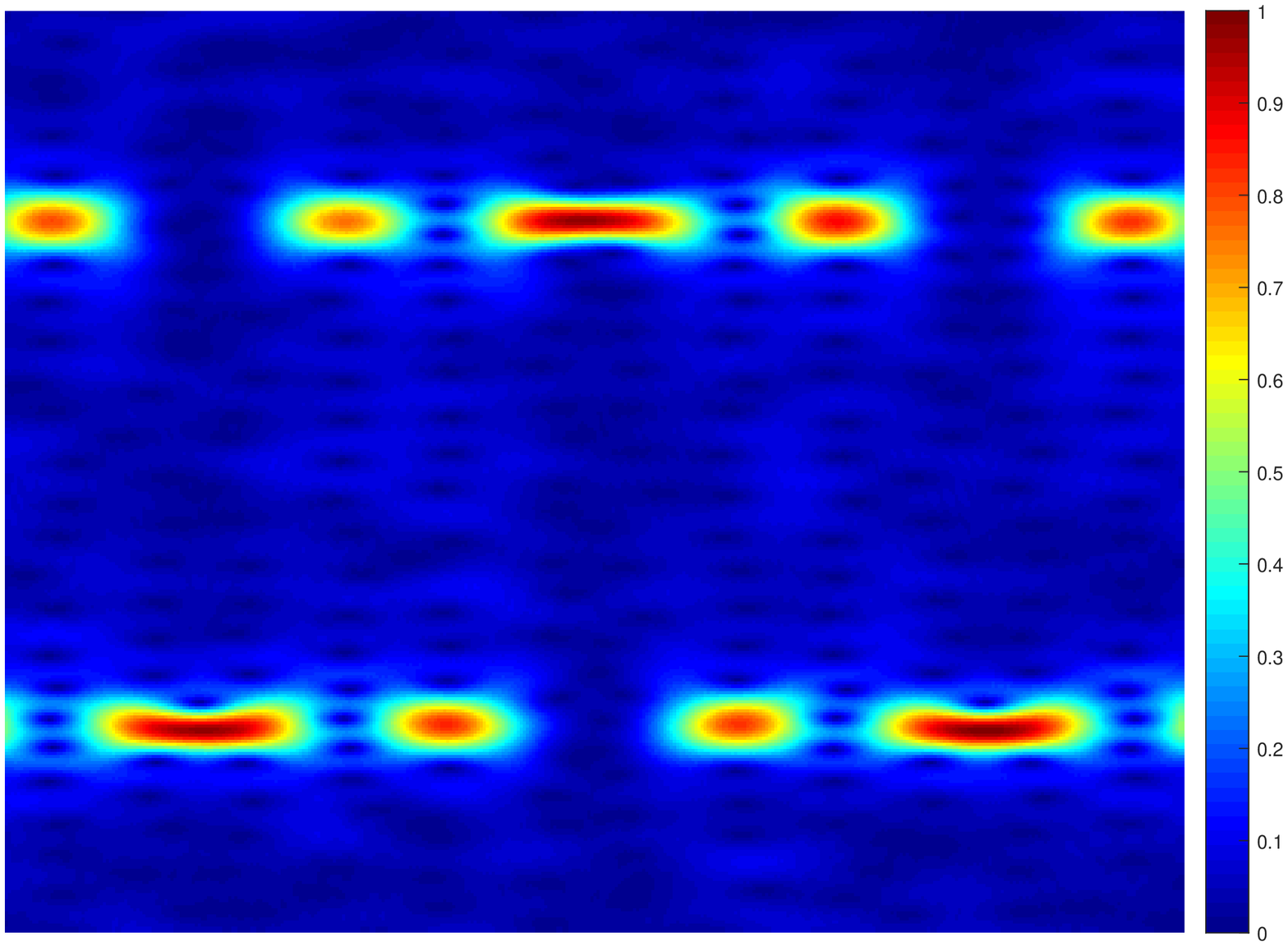} \\
\label{2FSK10dB}
\end{minipage}
}
\subfigure[2PSK]{
\begin{minipage}[b]{0.31\textwidth}
\includegraphics[width=\textwidth]{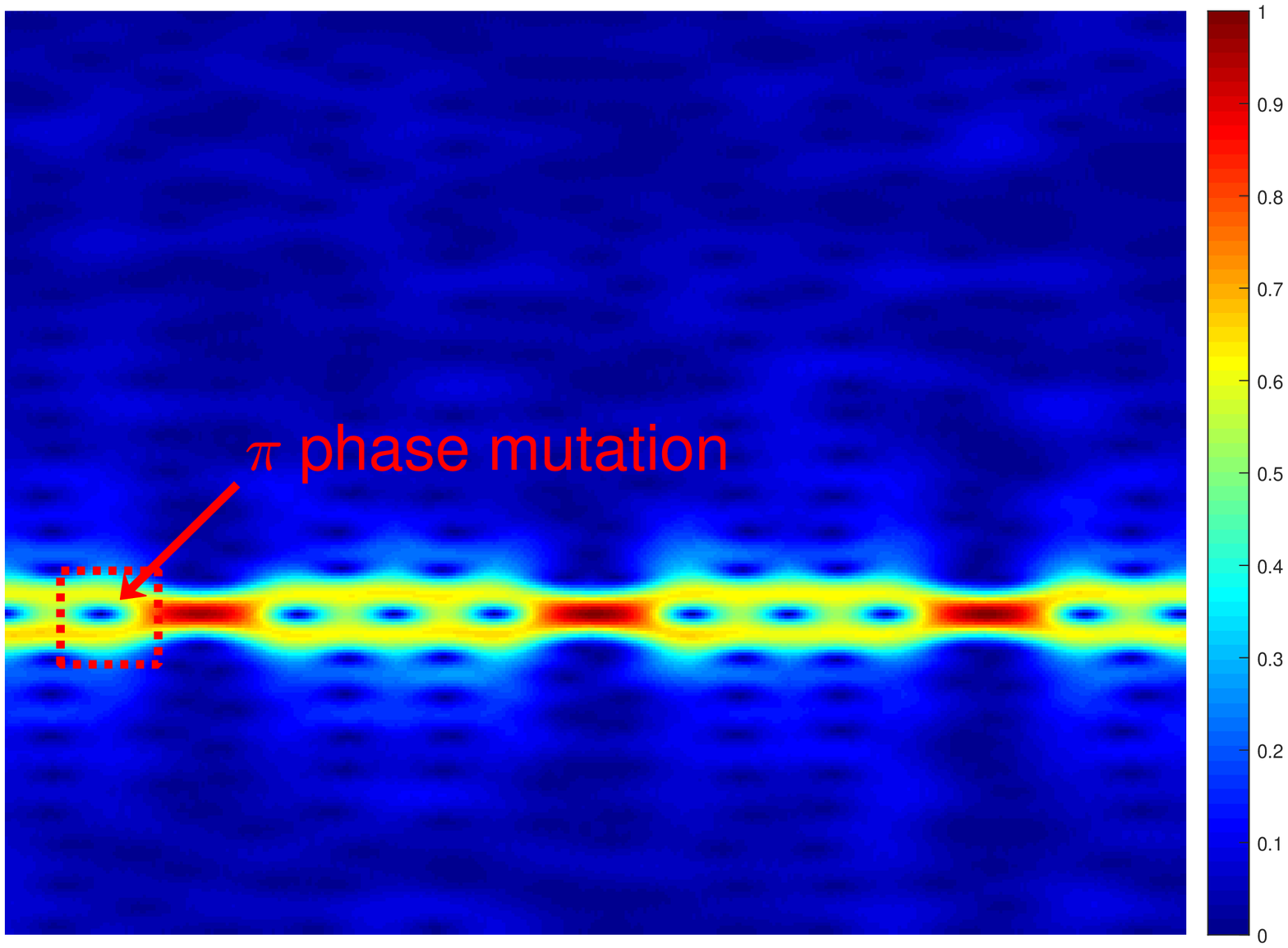} \\
\label{2PSK10dB}
\end{minipage}
}
\subfigure[4ASK]{
\begin{minipage}[b]{0.31\textwidth}
\includegraphics[width=\textwidth]{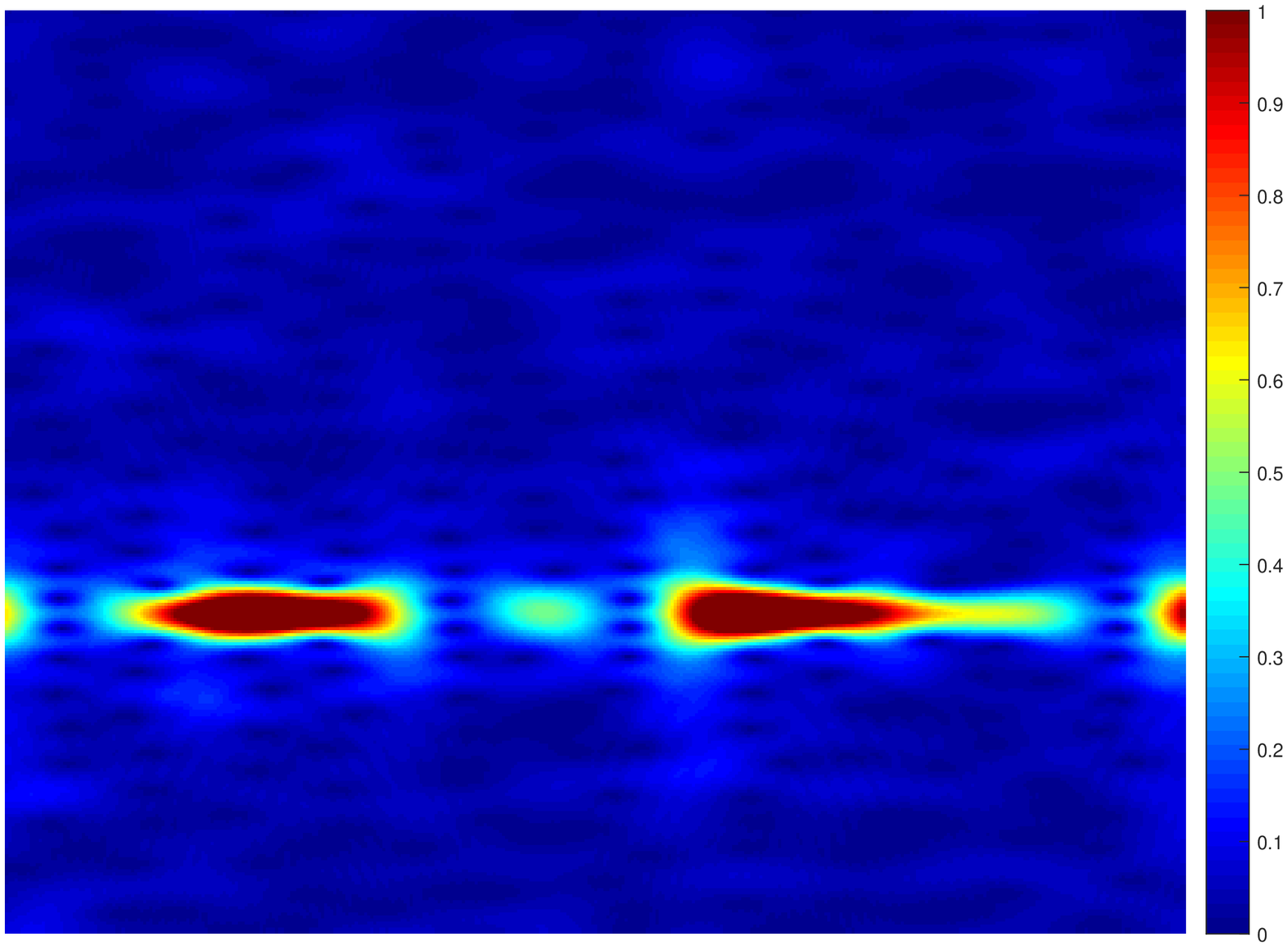} \\
\label{4ASK10dB}
\end{minipage}
}
\subfigure[4FSK]{
\begin{minipage}[b]{0.31\textwidth}
\includegraphics[width=\textwidth]{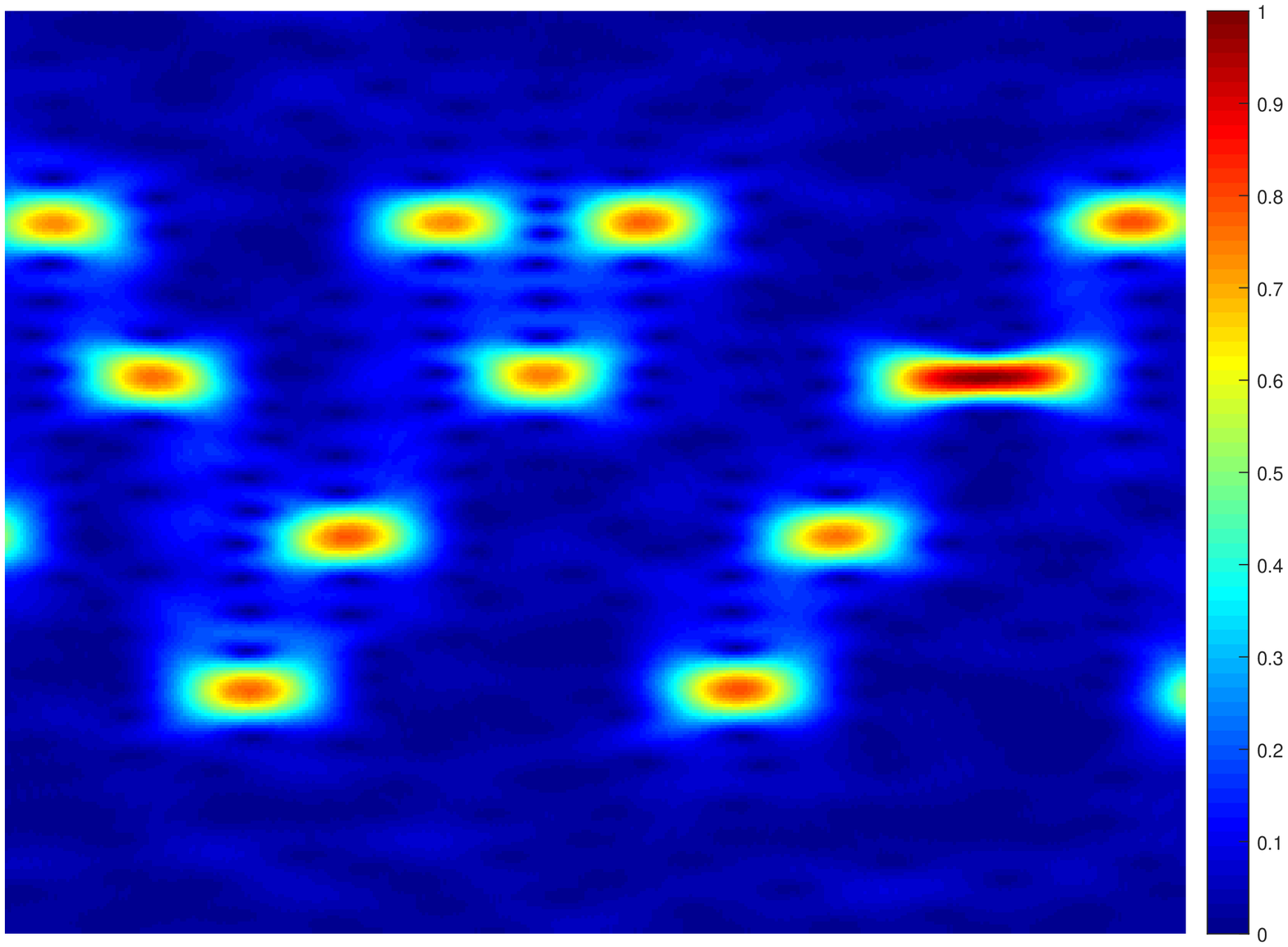} \\
\label{4FSK10dB}
\end{minipage}
}
\subfigure[4PSK]{
\begin{minipage}[b]{0.31\textwidth}
\includegraphics[width=\textwidth]{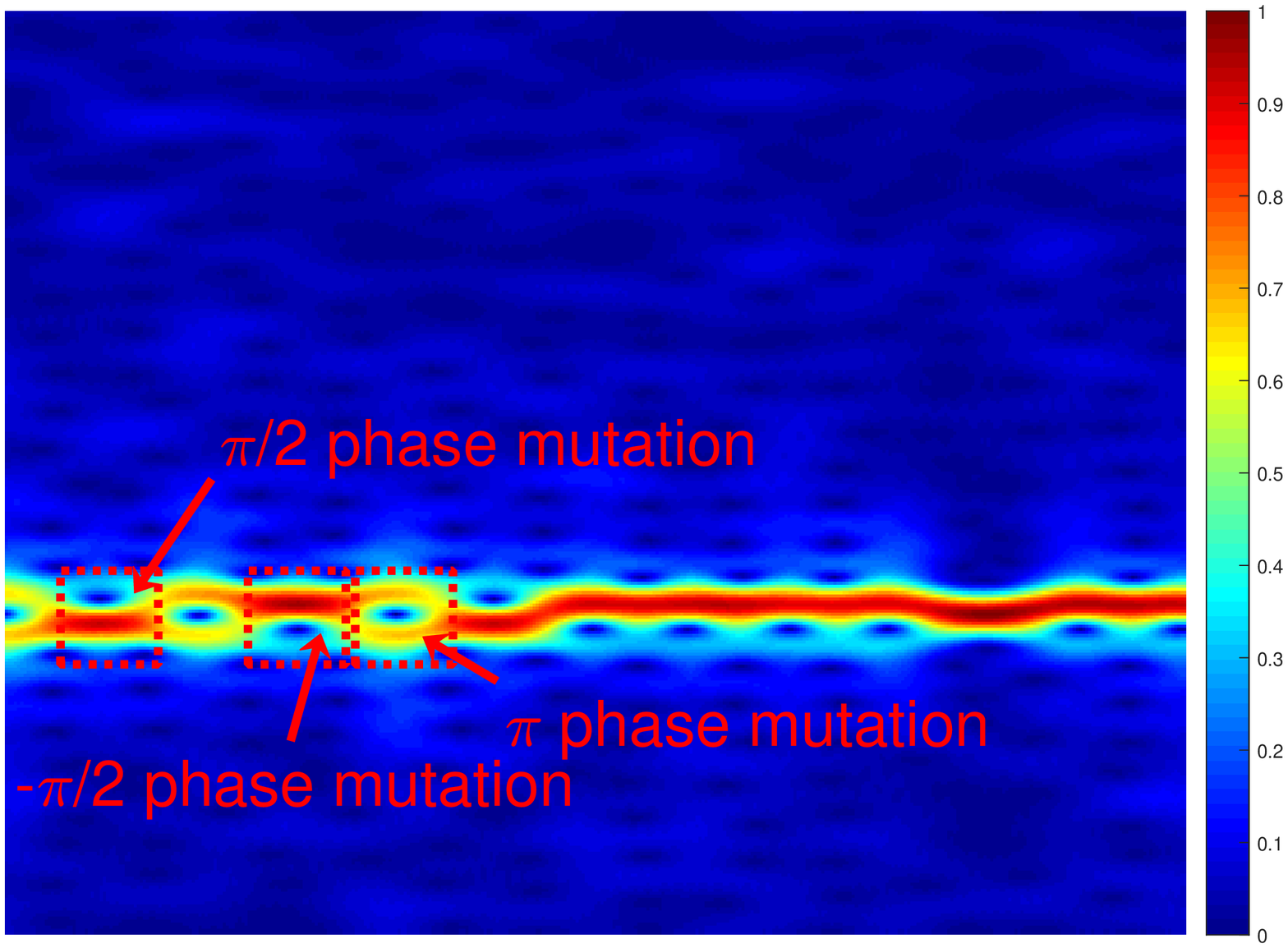} \\
\label{4PSK10dB}
\end{minipage}
}
\caption{$~$RGB spectrogram image of modulated signals in SISO networks at SNR = 10 dB.}
\label{RGBmodulation}
\end{figure*}

We first evaluate how the modulation order affects the RGB spectrogram image at a SNR of 10 dB for the SISO scenario. The considered modulation schemes are ASK, FSK, and PSK, which are shown in Fig. \ref{RGBmodulation}. They are analysed separately as follows.

First of all, the RGB spectrogram image is a time-frequency distribution image of the modulated signal. The horizontal axis of this image represents time and the vertical axis represents frequency. In addition, the colour of the RGB spectrogram image represents the value of the normalised spectral magnitude (i.e. the values corresponding to blue and red are zero and one, respectively).

Figs.  \ref{2ASK10dB} and \ref{4ASK10dB} show the RGB spectrogram image of the ASK-modulated signals. The power of the ASK-modulated signals concentrate on one frequency band in the image, and the power in the image is discontinuous over time. In addition, the colour in the image is blue when the digital signal sequence is at zero level in Fig. \ref{sequence}, and it is red when the digital signal sequence is at a non-zero level, which corresponds to the values of the spectral magnitude. In addition, compared with the 2ASK signal, the spectral magnitude of the 4ASK signal has a larger average value (i.e. more pixels in the 4ASK RGB spectrogram image have a value of 1).

\begin{figure*}
\centering
\subfigure[2ASK]{
\begin{minipage}[b]{0.31\textwidth}
\includegraphics[width=\textwidth]{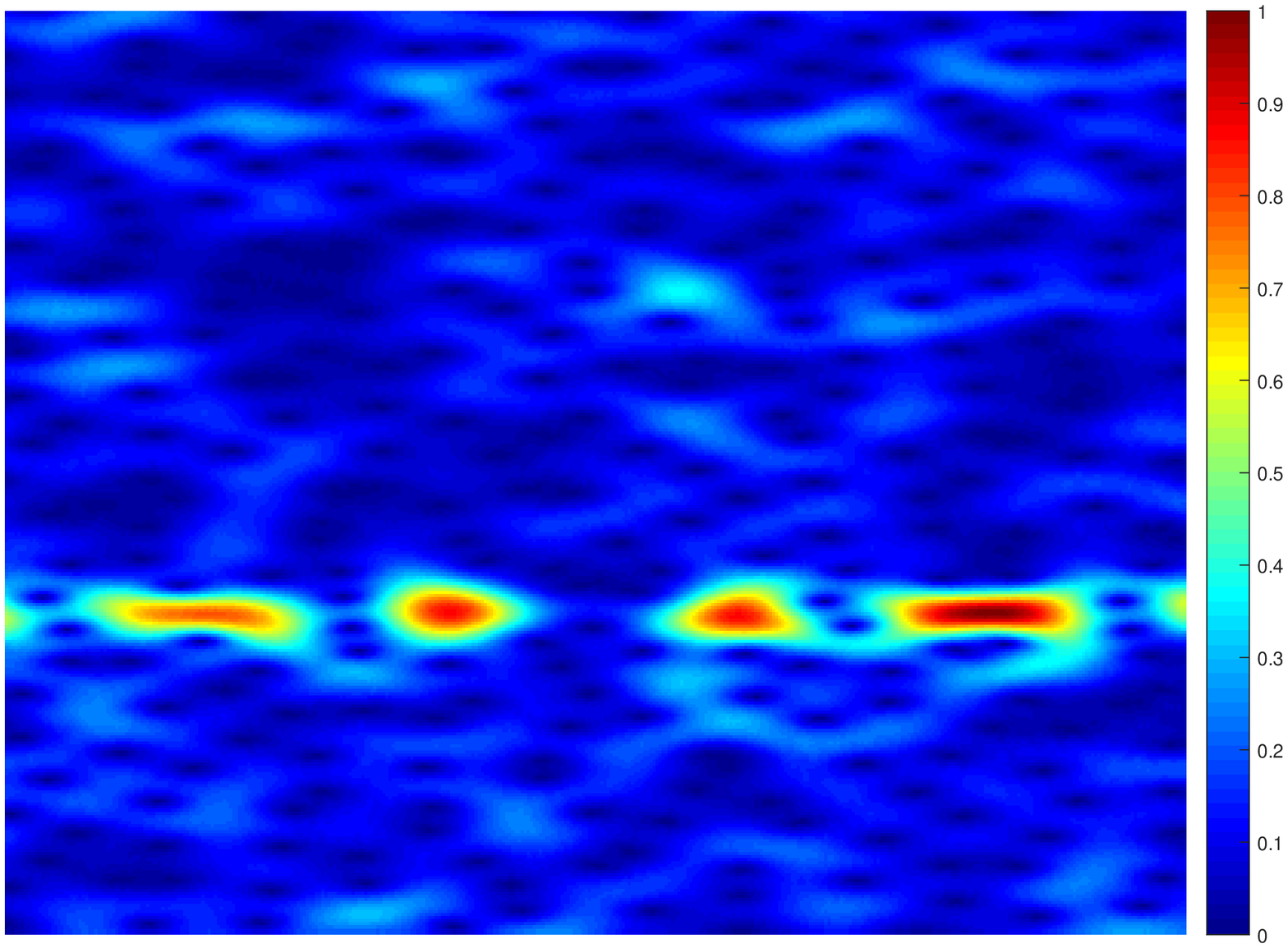} \\
\label{2ASKf4}
\end{minipage}
}
\subfigure[2FSK]{
\begin{minipage}[b]{0.31\textwidth}
\includegraphics[width=\textwidth]{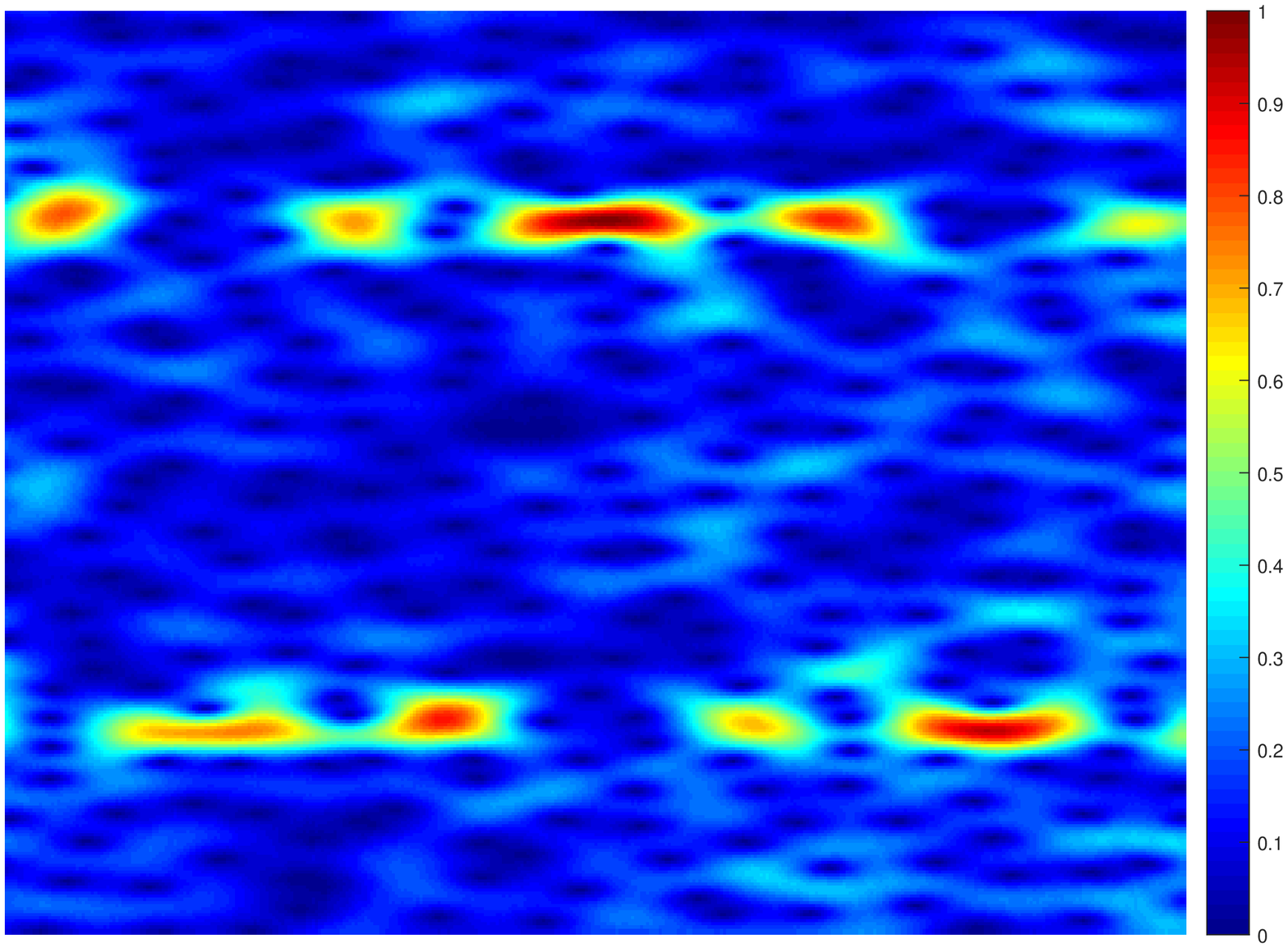} \\
\label{2FSKf4}
\end{minipage}
}
\subfigure[2PSK]{
\begin{minipage}[b]{0.31\textwidth}
\includegraphics[width=\textwidth]{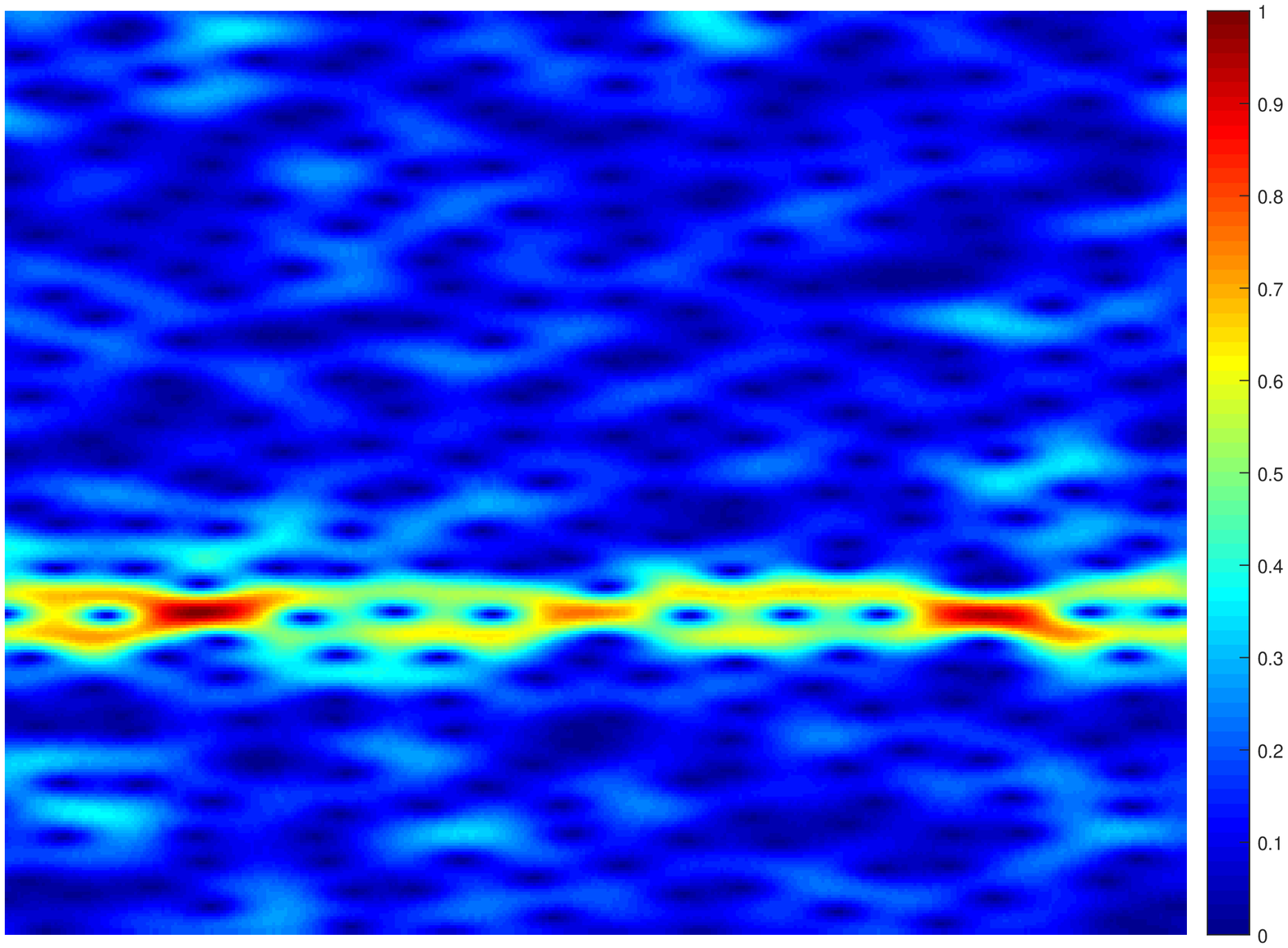} \\
\label{2PSKf4}
\end{minipage}
}
\caption{$~$RGB spectrogram image of modulated signals in SISO networks at SNR = -4 dB.}
\label{RGBf4}
\end{figure*}

Figs. \ref{2FSK10dB} and \ref{4FSK10dB} show the RGB spectrogram image of the FSK-modulated signals at a SNR of 10 dB. The spectral magnitude of the 2FSK modulated signals has a larger value over two subbands, and the spectral magnitude of the 4FSK modulated signals has a larger value over four subbands . For the FSK signals, the modulation order is equal to the number of modulated frequencies, which is the number of subbands in the RGB spectrogram image.

\begin{figure*}
\centering
\subfigure[2ASK]{
\begin{minipage}[b]{0.31\textwidth}
\includegraphics[width=\textwidth]{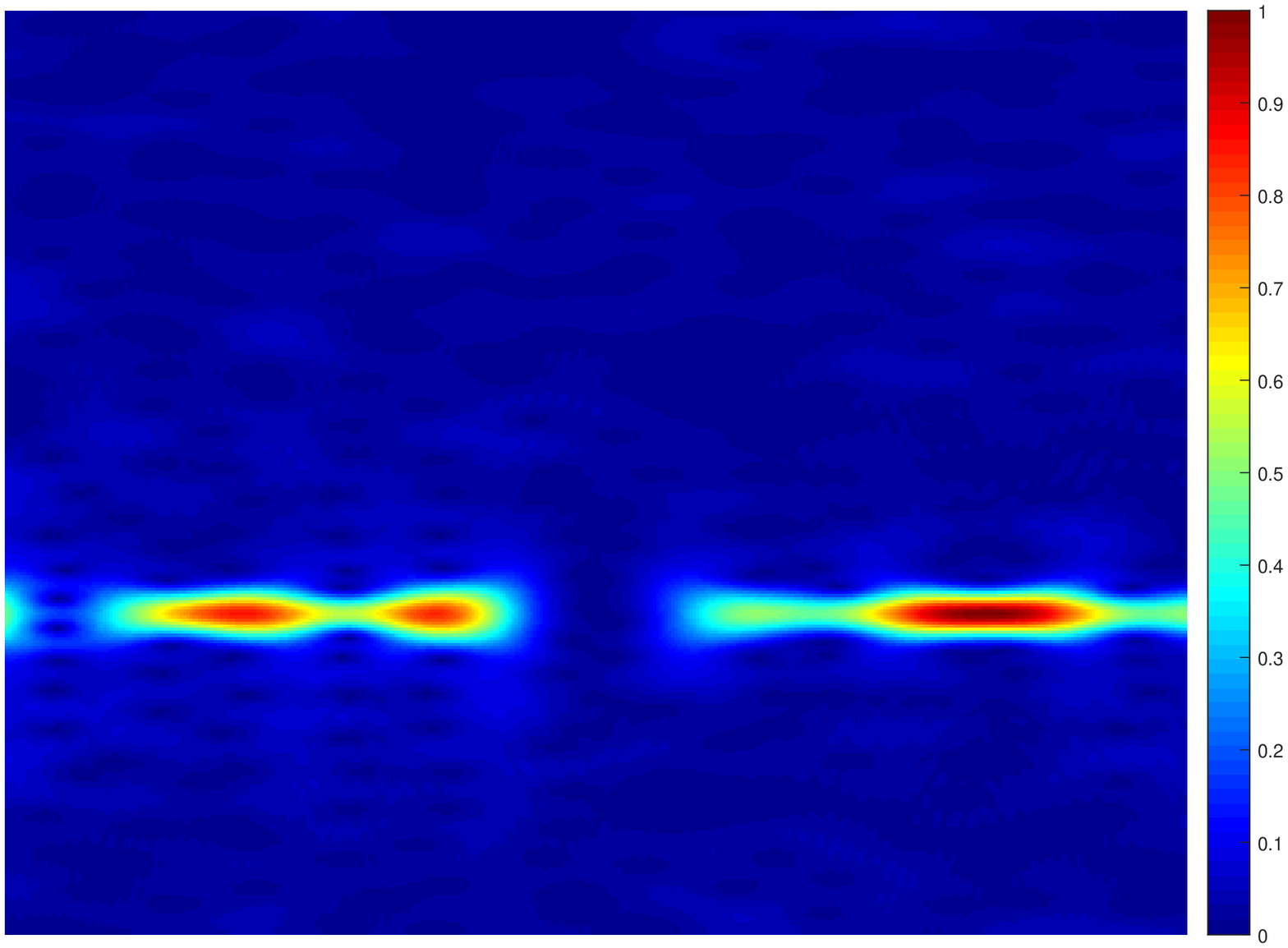} \\
\label{2ASKMIMO}
\end{minipage}
}
\subfigure[2FSK]{
\begin{minipage}[b]{0.31\textwidth}
\includegraphics[width=\textwidth]{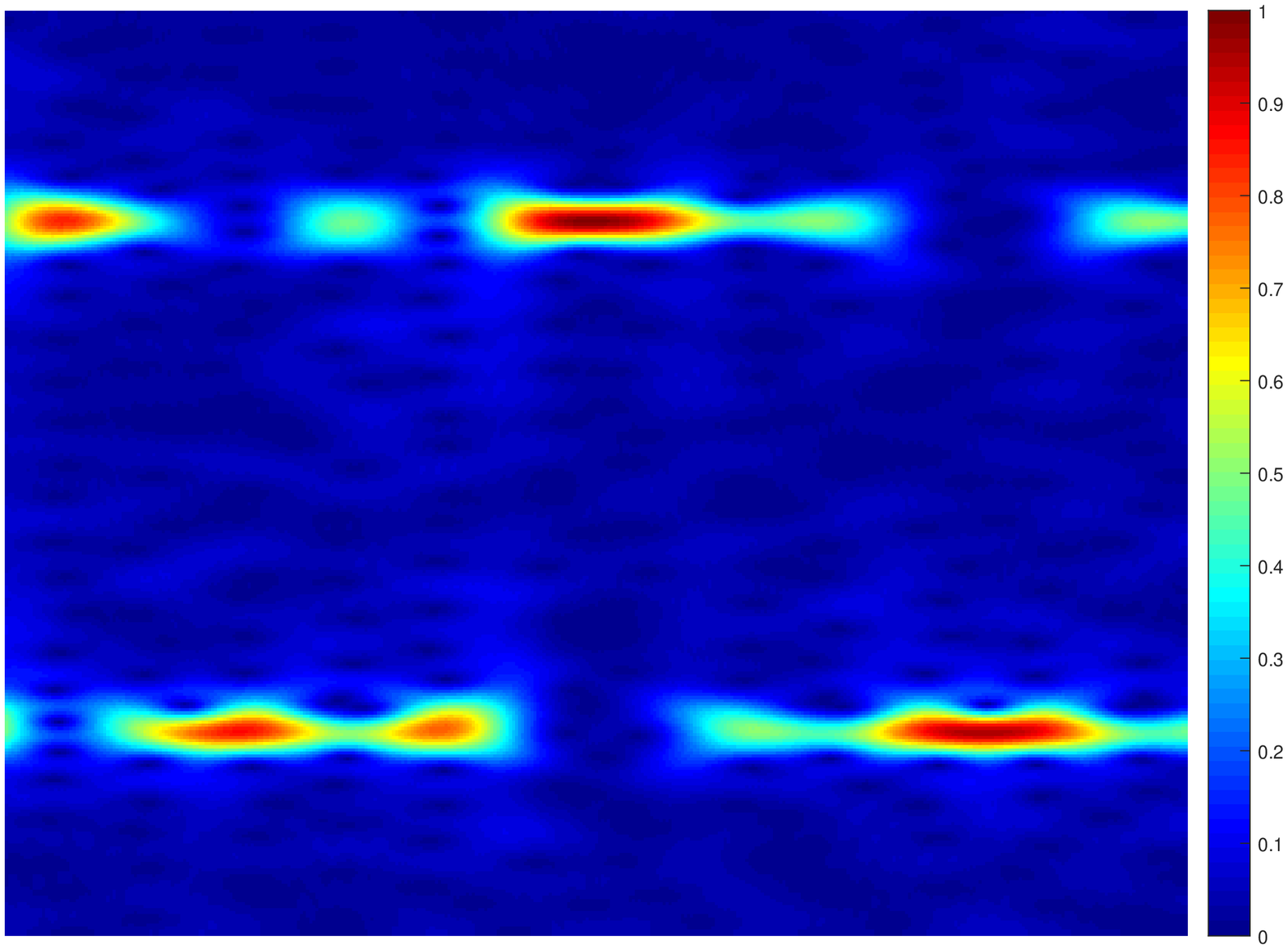} \\
\label{2FSKMIMO}
\end{minipage}
}
\subfigure[2PSK]{
\begin{minipage}[b]{0.31\textwidth}
\includegraphics[width=\textwidth]{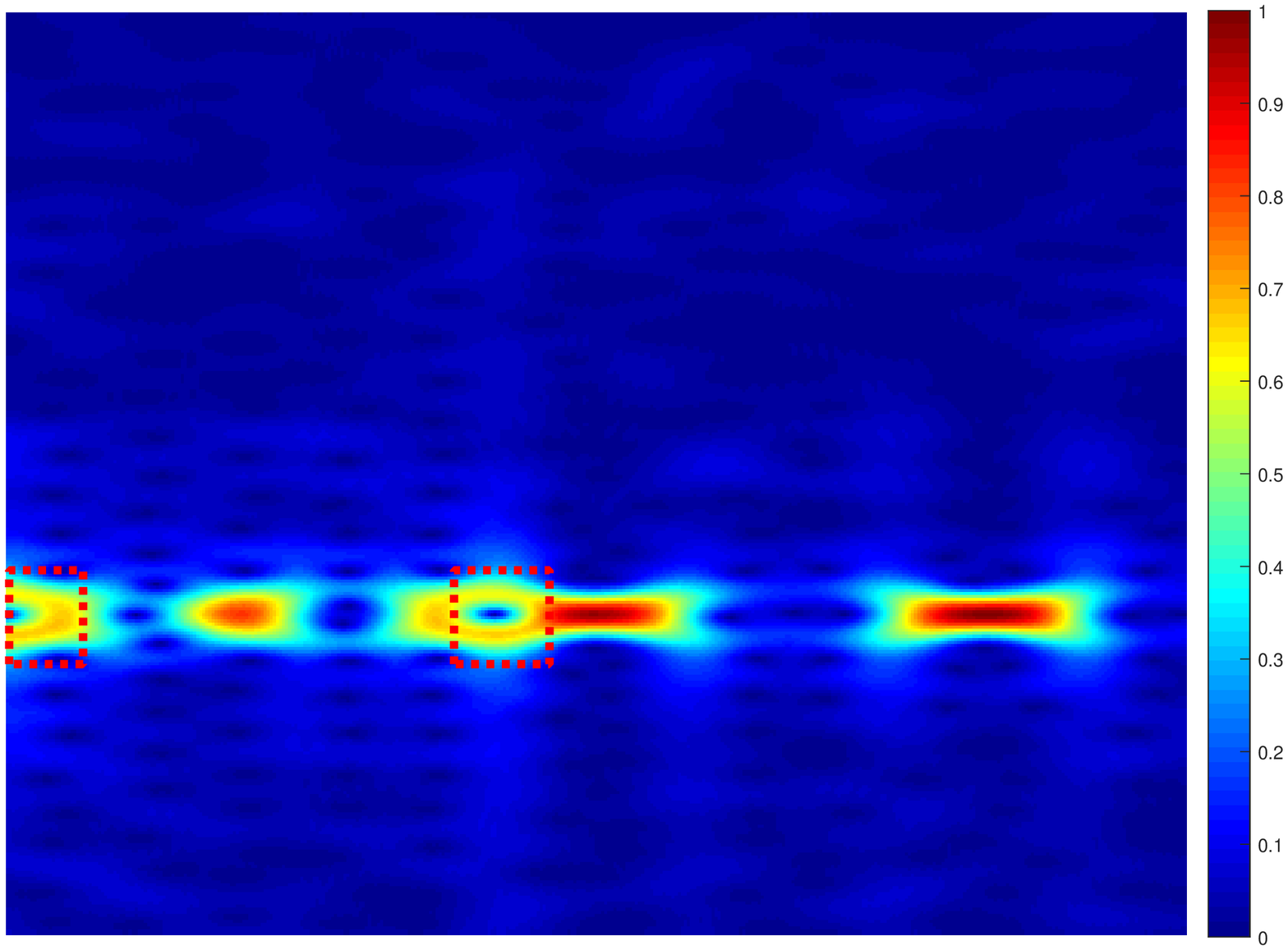} \\
\label{2PSKMIMO}
\end{minipage}
}
\caption{$~$RGB spectrogram image of modulated signals in MIMO networks at SNR = 10 dB.}
\label{RGBMIMO}
\end{figure*}

The RGB spectrogram images of the PSK-modulated signals are shown in Figs. \ref{2PSK10dB} and \ref{4PSK10dB}. The phase mutation of modulated signals is captured in the RGB spectrogram images. Specifically, Figs. \ref{sequencea} and \ref{2PSK10dB} both have the $\pi$ phase mutation in the 2PSK modulated signal from 0 to 1 and from 1 to 0 in the binary digital signal sequences. The $\pi$ phase mutation decreases the value of the power spectral density at the modulated frequency, which appears as a 'ring' in the RGB spectrogram image. Similarly, comparing Figs. \ref{sequenceb} and \ref{4PSK10dB}, the $\pi/2$ and $3\pi/2$ phase mutations also partly decrease the value of the power spectral density at the modulated frequency, but they appear as a 'half-ring' in the RGB spectrogram image. Therefore, modulated signals with different modulation orders have different time-frequency features, and it is reasonable to classify the modulated signals using the time-frequency analysis.

\subsubsection{RGB spectrogram image of the modulated signals with different SNRs}
In this paper, only the 2-order modulation schemes are analysed for different SNRs of the RGB spectrogram image. For the 2ASK modulated signals with $SNR=10$ dB and $SNR=-4$ dB, the corresponding RGB spectrograms are shown in Figs. \ref{2ASK10dB} and \ref{2ASKf4}, respectively. For the 2ASK-modulated signals, as the noise power increases, the components of the noise power become more prominent, as shown by the white patches in the RGB spectrogram image. However, the main features of the RGB spectrogram image of the 2ASK modulated signals are not destroyed. That is, the power distribution of the 2ASK-modulated signals is still concentrated in one subband in the RGB spectrogram image. In addition, the distribution of the power values of the power spectral density are almost the same at different SNRs. Similarly, the RGB spectrograms for the 2FSK- and 2PSK-modulated signals with $SNR=10$ dB and $SNR=-4$ dB are shown in Figs. \ref{2FSK10dB} and \ref{2FSKf4} and Figs. \ref{2PSK10dB} and \ref{2PSKf4}, respectively. From these figures, we can conclude that increases in the noise power do not destroy the main features of the RGB spectrogram image of these modulated signals, and thus they can be used as the features for modulation classification even in the low SNR region.

\subsubsection{RGB spectrogram image of the modulated signals for the MIMO channels}
We now analyse how the MIMO channel influences the RGB spectrogram image of the modulated signals. The 2ASK, 2FSK, and 2PSK modulation schemes are discussed herein. The antenna configuration for the MIMO system is $N_t=2$ and $N_r=4$, then the random channel attenuation assigns a value from $[0,1]$, and random phase shifts within one symbol interval are considered for the MIMO scenario, and the  AWGNs with 10dB SNRs are added into the modulated signals. In addition, a multiplexing-based transmission scheme is adopted for the MIMO system. Specifically, two transmit antennas send two independent data streams, but with the same modulation scheme (e.g. 2ASK, 2FSK, or 2PSK). The result is shown in Fig. \ref{RGBMIMO}.

A comparison of Figs. \ref{RGBMIMO} and \ref{RGBmodulation} shows that, for all the modulated signals, the signal overlapping of the MIMO system has no effect on the power distribution of the modulated signals in the frequency domain, but the power distribution over the time domain is changed. The latter can be explained by the fact that the overlapping of different transmitted signals partly destroys the time-frequency characteristics of raw modulated signals. In spite of this, some crucial time-frequency characteristics are not destroyed by the MIMO signal overlapping, such as the 'ring' that is caused by the phase mutation in the 2PSK signal (shown in Figs. \ref{2PSK10dB} and \ref{2PSKMIMO}). Hence, the overlapping of modulated signals partially destroy the time-frequency characteristics, but some of the crucial time-frequency characteristics are still preserved in the RGB spectrogram image. Therefore, the RGB spectrogram image can still be used to identify the modulation type, even in the MIMO scenario.
\begin{figure}[htbp]
\centerline{\includegraphics[scale=0.38]{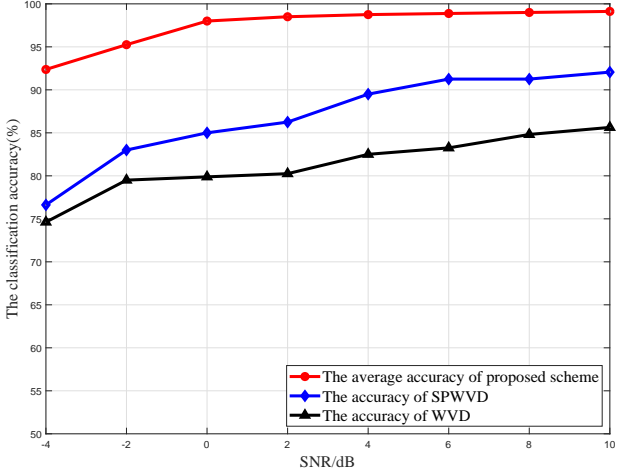}}
\caption{$~$Classification accuracy versus SNR in SISO networks.}
\label{SISOaccuracy}
\end{figure}

\subsection{Classification accuracy of proposed scheme}
\label{Accuracy}
The classification accuracy of the proposed scheme is tested and verified for both the SISO and MIMO scenarios. We first randomly generate the data stream, and then it is modulated and passed through the SISO or MIMO channels. In order to verify the performance of the proposed scheme, some benchmark schemes are introduced, such as the scheme based on the smooth pseudo Wigner-Ville distribution (SPWVD) proposed in \cite{8643801} and the scheme based on the Wigner-Ville distribution (WVD) proposed in \cite{2016xli}.

\begin{figure}[htp]
\centering
\subfigure[SNR = -4dB]{
\begin{minipage}[b]{0.5\textwidth}
\includegraphics[width=\textwidth]{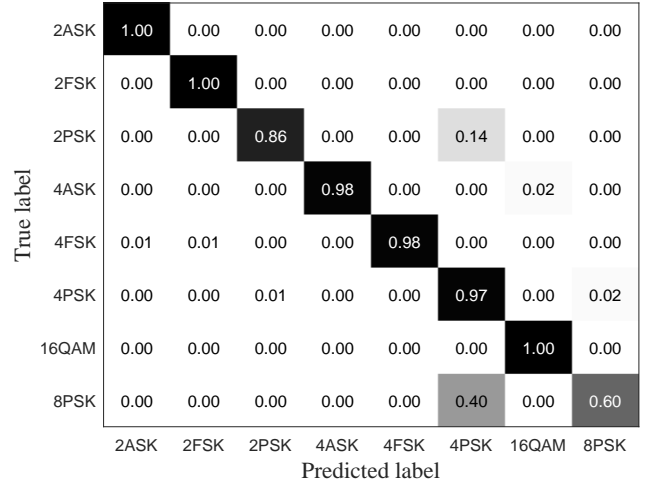} \\
\label{SISOf4dBmat}
\end{minipage}
}
\subfigure[SNR = 10dB]{
\begin{minipage}[b]{0.5\textwidth}
\includegraphics[width=\textwidth]{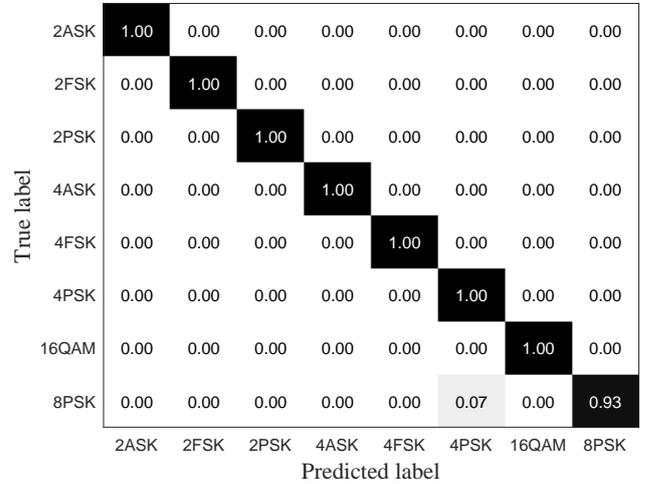} \\
\label{SISO10dBmat}
\end{minipage}
}
\caption{$~$Confusion matrices of proposed scheme in SISO networks.}
\label{SISOmat}
\end{figure}
\subsubsection{Classification accuracy in the SISO scenario}
For the SISO scenario, the average classification accuracy of the proposed scheme is evaluated by varying the SNR of the signals from -4 dB to 10 dB. The result is shown in Fig. \ref{SISOaccuracy}. As the SNRs of the signals increase, the classification accuracies of all three classification schemes  gradually improve. Moreover, our proposed scheme always has the highest average accuracy. Its classification accuracy is always larger than 92.37 \%, even at $SNR=-4$ dB, and it can reaches a classification accuracy of 99.12\% at $SNR=10$ dB. This significantly outperforms the SPWVD- and WVD-based methods.  These results show that our method has high accuracy and robustness, even at low SNR.

The confusion matrices of the classification results, from which the classification accuracy of each modulation type can be derived, are shown in Fig. \ref{SISOmat}. Figure \ref{SISOf4dBmat} indicates that if the SNR of the modulated signals is low, the classification accuracies of the 2PSK- and 8PSK-modulated signals are low. For example, at $SNR=-4$ dB, the 2PSK and 8PSK signals may be incorrectly identified as 4PSK with probabilities of 0.14 and 0.4, respectively. This can be explained by the fact that different PSK signals have similar time-frequency characteristics in the RGB spectrogram image, especially for the 4PSK- and 8PSK-modulated signals. By contrast, with a high SNR, all the modulated signals are successfully identified except the 8PSK signals, which achieved an identification accuracy of 93\%. Therefore, the proposed time-frequency analysis and deep learning-based BMC scheme can achieve excellent performance at both low and high SNR in the SISO scenario.

\subsubsection{Classification accuracy in the MIMO scenario}
\begin{figure}[htbp]
\centerline{\includegraphics[scale=0.38]{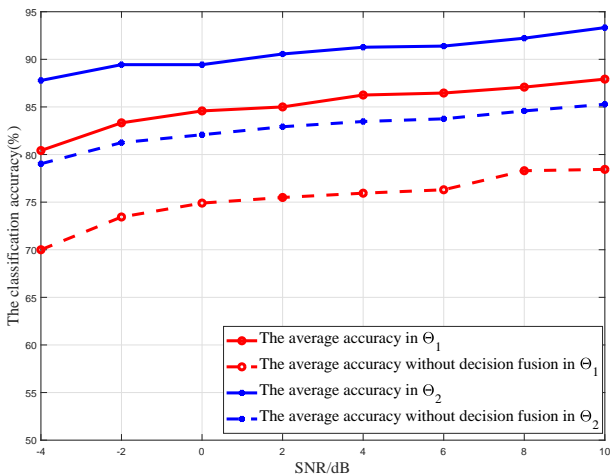}}
\caption{$~$Classification accuracy versus SNR in MIMO networks.}
\label{MIMOaccuracy}
\end{figure}

The classification performance of the proposed scheme in the MIMO scenario is now verified. In order to better understand the performance of the proposed scheme, the model is trained and tested with two data sets (as in \cite{6117042}): one for the modulation set $\mathbf{\Theta_1}$ = \{2ASK, 2FSK, 2PSK, 4ASK, 4FSK, 4PSK, 8PSK, 16QAM\} and another for a smaller modulation set $\mathbf{\Theta_2}$ = \{2ASK, 2FSK, 2PSK, 4ASK, 4FSK, 4PSK\}. In the testing stage , the SNR of the modulated signals is varied from $SNR=-4$ dB to $SNR=10$ dB, and the result is shown in Fig. \ref{MIMOaccuracy}. For both scenarios with and without the decision fusion module, the classification accuracy of the proposed scheme increases as the SNR of the modulated signals increases, which is consistent with the theoretical analysis. However, by introducing the decision fusion module, a 10\% performance improvement in the classification accuracy can be achieved. In more detail, the proposed scheme can achieve 80.42\% and 87.92\% accuracy at -4 and 10dB SNR in $\mathbf{\Theta_1}$, and 87.78\% and 93.33\% accuracy at -4 and 10dB SNR in $\mathbf{\Theta_2}$. In addition, the average classification accuracy for the MIMO scenario is lower than the SISO scenario. This is due to the fact that, by using multiple antennas in the system, the structure of the original signals is destroyed by overlapping at the receive antenna, as mentioned in section \ref{RGBDiscuss}.

\begin{figure}[htp]
\centering
\subfigure[SNR = -4dB]{
\begin{minipage}[b]{0.5\textwidth}
\includegraphics[width=\textwidth]{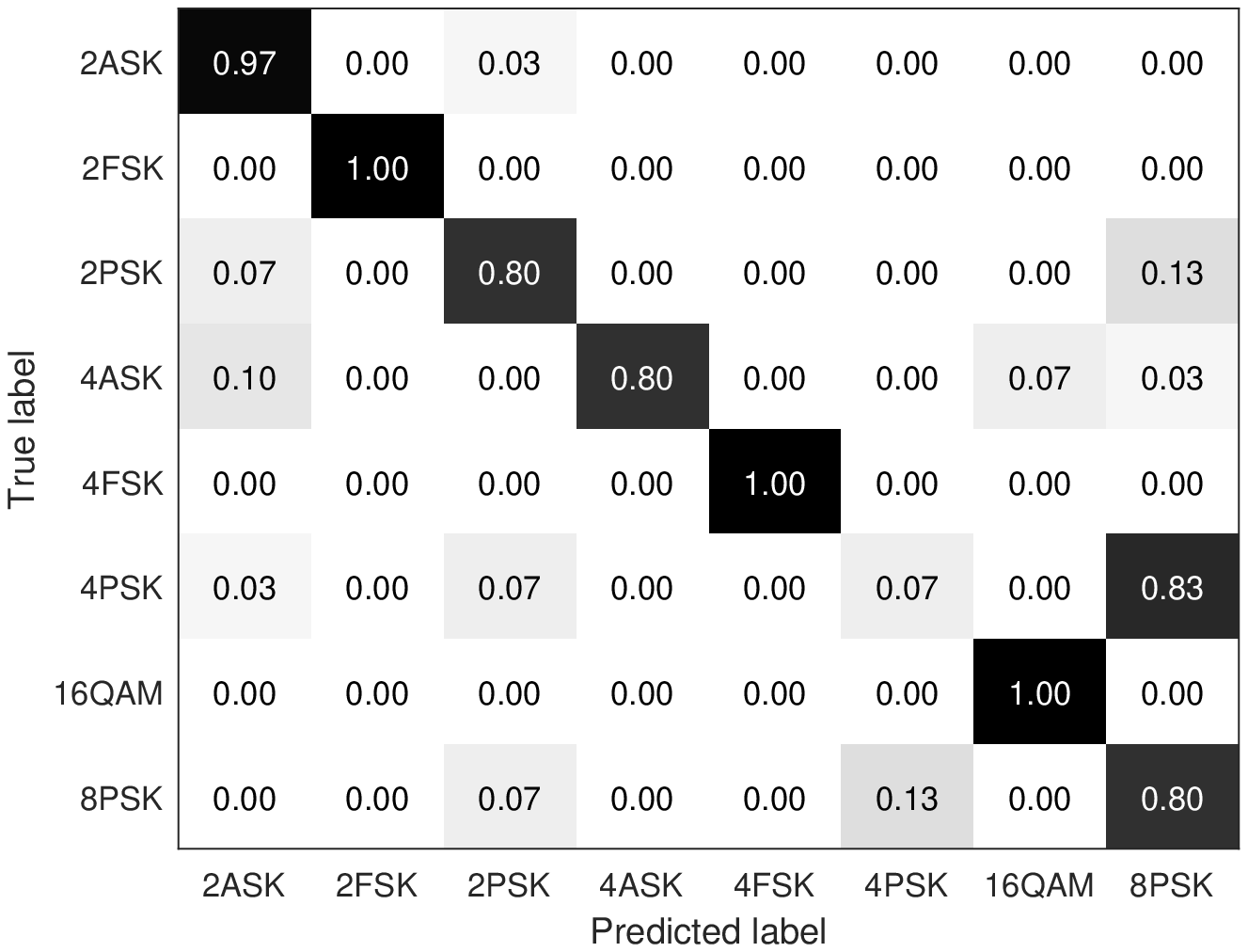} \\
\label{MIMOf4dBmat}
\end{minipage}
}
\subfigure[SNR = 10dB]{
\begin{minipage}[b]{0.5\textwidth}
\includegraphics[width=\textwidth]{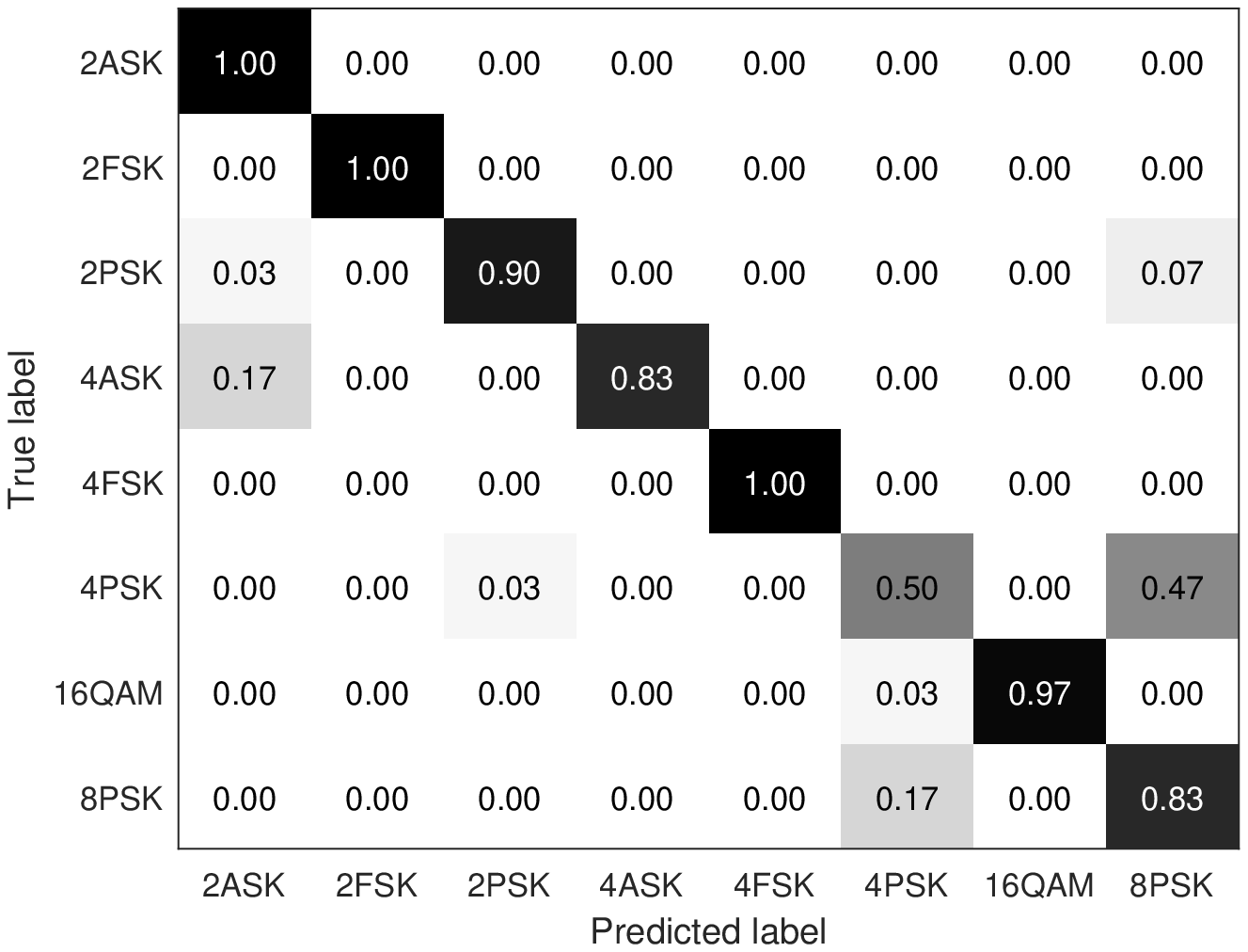} \\
\label{MIMO10dBmat}
\end{minipage}
}
\caption{$~$Confusion matrices of proposed scheme in MIMO networks.}
\label{MIMOmat}
\end{figure}

Similarly, the confusion matrices of the classification results are shown in Fig. \ref{MIMOmat}(a) and (b) for modulated signal SNRs of -4 dB and 10 dB, respectively. The MFSK- and QAM-modulated signals have the highest classification accuracies at both -4 dB and 10 dB, and the MASK-modulated signals have the second highest . The MPSK signals (especially the 4PSK signals) exhibit the worst classification performance, as shown in Fig. \ref{MIMOf4dBmat}. Most of the 4PSK are misclassified as 8PSK at $SNR=-4$ dB, and the performance is improved only slightly at $SNR=10$ dB. This result indicates that the MIMO system structure has negative effects on the time-frequency characteristics of the MPSK signals, which is consistent with the theoretical analysis. Hence, our proposed scheme has difficulty identifying the high-order PSK signals in the MIMO system. However, the time-frequency analysis and deep learning-based scheme have excellent performance in classifying the MFSK-, ASK-, and QAM-modulated signals, and it can obtain superior average classification accuracy for the MIMO system.

\section{Conclusion}
\label{conclusion}
In this paper, we resolve the problem of blind modulation classification for the MIMO system. Specifically, the windowed STFT was used to analyse the time-frequency characteristics of the modulation signals, and the time-frequency graphs of the modulated signals were converted to RGB spectrogram images. Then transfer learning was utilised to fine-tune AlexNet to adapt to our classification problem, and the generated RGB spectrogram images were fed into the fine-tuned CNN to extract features and train the net. Finally, the decision of each received signal from the MIMO receivers were combined by the decision fusion module for the final decision. The STFT-based time-frequency analysis results showed that each modulation type had unique time-frequency characteristics, and that the additive noise had limited influence on the time-frequency characteristics of the modulation signals. The final classification results indicated that the proposed scheme can achieve 92.37\% and 99.12\% classification accuracy at SNRs of -4 dB and 10 dB in the SISO scenario. For the MIMO system, the proposed scheme still achieved 70\% and 80\%  at -4 dB for the large and small modulation sets, respectively. In future work, we plan to improve the performance of the proposed scheme for the high-order PSK signals.

\bibliography{bibfile}

\end{document}